\documentclass{article}

\usepackage[english]{babel}

\usepackage[letterpaper,top=2cm,bottom=2cm,left=3cm,right=3cm,marginparwidth=1.75cm]{geometry}

\usepackage[utf8]{inputenc} % allow utf-8 input
\usepackage[T1]{fontenc}    % use 8-bit T1 fonts
\usepackage{url}            % simple URL typesetting
\usepackage[round,authoryear]{natbib}
\usepackage{booktabs}       % professional-quality tables
\usepackage{amsfonts}       % blackboard math symbols
\usepackage{nicefrac}       % compact symbols for 1/2, etc.
\usepackage{microtype}      % microtypography
\usepackage{xcolor}         % colors
\usepackage{amsmath}
\usepackage{graphicx}
\usepackage{times}
\usepackage{soul}
\usepackage{xspace}
\usepackage[small]{caption}
\usepackage{placeins}
\usepackage{amsthm}
\usepackage{algorithm}
\usepackage{algorithmic}
\usepackage{amssymb}
\usepackage{multirow}
\usepackage{array}
\usepackage[
  unicode=true,
  bookmarks=true,
  bookmarksnumbered=true,
  colorlinks=true,
  linkcolor=blue!55!black,
  citecolor=blue!55!black,
  urlcolor=blue!55!black
]{hyperref}

\hypersetup{
  bookmarksopen=true,
  pdfstartview={FitH},
  pdfborder={0 0 0}
}
\newcolumntype{P}[1]{>{\raggedright\arraybackslash}p{#1}}
\title{Beyond Spatial Compression: Interface-Centric Generative States for Open-World 3D Structure}
\author{
Xiang Chen$^{1}$ \quad Alexander Binder$^{1,2,3}$\\
$^{1}$DSC ScaDS.AI, Leipzig University\\
$^{2}$Institute for Cancer Genetics and Informatics (ICGI), Oslo, Norway\\
$^{3}$ICT Cluster, Singapore Institute of Technology, Singapore\\
\texttt{\{xiang.chen, alexander.binder\}@uni-leipzig.de}
}

\begin{document}
\maketitle

\begin{abstract}
Current 3D tokenizers largely treat representation as spatial compression: compact codes reconstruct surface geometry, but leave component ownership and attachment validity implicit. In open-world assets with intersecting components, noisy topology, and weak canonical structure, this creates a representation mismatch: local shape, component identity, and assembly relations become entangled in a latent stream and are not natively addressable during decoding. We formulate an alternative view, \emph{interface-centric generative states}, in which tokenization constructs an operational state rather than a passive compressed code. The state exposes local geometry, component ownership, and attachment validity as variables that can be queried, constrained, and repaired during decoding.
We instantiate this formulation with Component-Conditioned Canonical Local Tokens (C2LT-3D), factorizing representation into canonical local geometry, partition-conditioned context, and relational seam variables. Each factor targets a distinct failure mode of compression-centric tokens: pose leakage, cross-component interference, or invalid local attachment. This exposed state supports attachment validation, latent structural repair, targeted intervention, and constrained serialization without a separate post-hoc structure recovery module. Trained on single-object CAD models and evaluated zero-shot on open-world multi-component assets, C2LT-3D improves structural robustness and shows that its latent variables remain actionable under adversarial attachment settings. These results suggest that open-world 3D generative representations should be evaluated not only by reconstruction fidelity, but by whether their discrete states remain operational for assembly-level structural reasoning.

\end{abstract}

% !TEX root =  neurips_2026.tex

% =========================================================================
% 1. INTRODUCTION
% =========================================================================
\section{Introduction}
\label{sec:intro}

Discrete tokenization has transformed language because tokens are not merely compressed observations; they are units over which syntax, semantics, and reasoning can be expressed. In 3D generation, however, tokenization is still largely treated as \emph{spatial compression}: recent mesh and surface tokenizers \citep{polygen, meshgpt, bpt, lost} map geometry to short discrete sequences for autoregressive modeling. This works well on clean, single-component CAD objects such as ShapeNet \citep{shapenet}, but is less reliable on open-world assets from Objaverse and Objaverse-LVIS \citep{objaverse, objaverse_lvis}, where objects contain intersecting components, non-manifold interiors, arbitrary scale, and weak canonical orientation.

\begin{figure}[t]
\centering
\includegraphics[width=0.97\textwidth]{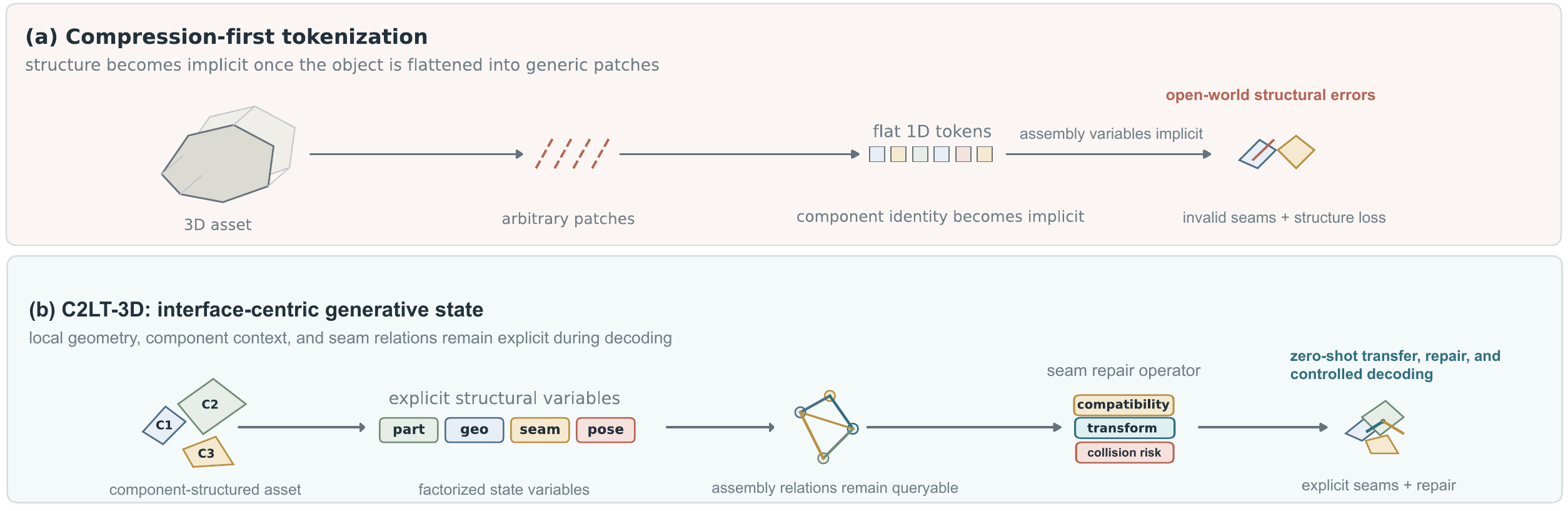}
\caption{\textbf{From Compression Tokens to Interface-Centric Generative States.} (a) Compression-first tokenization flattens complex 3D assets into generic patch sequences, leaving assembly variables implicit and exposing open-world assets to structural errors. (b) C2LT-3D exposes local geometry, component context, and seam relations as a generative state rather than a compressed token stream, enabling object-level realization, latent repair, structural intervention, and controlled decoding.}
\label{fig:teaser}
\end{figure}

We argue that this limitation reflects a representation mismatch. Compression-centric tokens treat an asset as arbitrarily divisible geometry, while many open-world objects behave more like assemblies whose local shape, component identity, and attachment structure must remain distinguishable. We therefore propose an \emph{interface-centric generative-state paradigm for open-world 3D structure} (Figure~\ref{fig:teaser}): a model should not only predict compressed geometry tokens, but construct an operational structural state in which geometry, ownership, and attachment relations remain explicit throughout decoding. We use ``paradigm'' operationally: a generative representation defines a paradigm when it determines not only how objects are encoded, but what operations are available during generation. Compression-centric tokens primarily support reconstruction from a passive latent stream; interface-centric generative states additionally support attachment validation, latent repair, intervention, and constraint-aware decoding. Thus tokenization becomes a choice of state variables, and those variables determine which structural operations are native rather than recovered after decoding.

Our central hypothesis is that open-world structural errors arise because intrinsic local shape, macro-component membership, and local assembly are entangled inside compression-centric tokens. We introduce Component-Conditioned Canonical Local Tokens (\textbf{C2LT-3D}), an interface-centric generative-state representation that separates these variables without semantic part annotations. The state combines \textbf{Canonical Local Geometry}, which stabilizes local frames so the codebook captures shape rather than pose variants; \textbf{Partition-Conditioned Context}, which injects unsupervised macro-component ownership as a soft structural bias; and a \textbf{Relational Seam Prior for Controlled Decoding}, which scores physical compatibility, refines relative transforms, and predicts collision risk between adjacent tokens. A deterministic \textbf{Component-Owned Object Realization} map resolves the local-to-global decode mismatch by suppressing cross-component leakage during object reconstruction and mesh visualization.

\textbf{Contributions and scope.}
We formulate open-world 3D tokenization as construction of an \emph{operational generative state}, identifying a mismatch between geometry fidelity and assembly-level validity. C2LT-3D instantiates this state with a compact factorization into canonical local geometry, partition-conditioned context, and relational seams; each factor targets a distinct failure mode of compression-centric tokens. We then show that the resulting variables are addressable for attachment validation, latent repair, structural intervention, and constrained decoding without a separate post-hoc structure recovery module. The goal is not unconditional generation alone, but whether tokens expose repairable assembly-level variables. We therefore evaluate C2LT-3D through zero-shot transfer to open-world multi-component assets, latent-space repair and targeted intervention, and controlled serialization under a fixed lightweight decoder. Across these settings, C2LT-3D supports the view that open-world 3D generative states should be judged not only by surface fidelity, but by whether their discrete variables remain actionable for assembly-level reasoning.

% !TEX root =  neurips_2026.tex

% =========================================================================
% 2. RELATED WORKS
% =========================================================================
\section{Related Work}
\label{sec:related}

\textbf{Compression-centric 3D tokenization.}
Discrete latent-variable models \citep{vqvae, vqgan} are now central to generative modeling. In 3D, PolyGen \citep{polygen}, MeshGPT \citep{meshgpt}, BPT \citep{bpt}, LoST \citep{lost}, MeshAnythingV2 \citep{meshanything}, and MeshArt \citep{meshart} all reduce geometry to compact token sequences or local blocks for efficient generation. These methods are powerful on clean CAD domains, but their compression objective makes component identity and attachment structure implicit. C2LT-3D instead treats tokenization as construction of an interface-centric generative state: local shape, component ownership, and seam compatibility are represented separately so they can be evaluated and repaired.

\textbf{Structural and relational 3D representations.}
Part-based methods based on PartNet \citep{partnet}, StructureNet \citep{structurenet}, BSP-Net \citep{bspnet}, or direct geometric proxy fitting \citep{neuralparts} demonstrate the value of compositional structure, but often require semantic annotations or category-specific assumptions. LLM-assisted mesh reasoning \citep{meshllm, llamamesh} offers another route but delegates structure to external language priors. Our goal is different: C2LT-3D learns from geometry alone, using unsupervised partitions and a seam prior to expose a discrete generative state for attachment ranking, latent repair, and structural intervention.

\textbf{Canonicalization and assembly constraints.}
Canonical frames can stabilize 3D learning \citep{pointnet, deepsdf}, but purely global or PCA-based local frames are fragile for symmetric, planar, or noisy patches. We use partial canonicalization with explicit fallbacks to improve token reuse, and we embed local assembly constraints into the latent state in the spirit of computational fabrication \citep{mitra2014computational}. We do not claim that semantic concepts such as wheels or supports automatically emerge; rather, we provide an operational state on which geometric and relational structure can be measured directly.

% !TEX root =  neurips_2026.tex

% =========================================================================
% 2. PROBLEM SETUP
% =========================================================================
\section{An Interface-Centric Generative State Space}
\label{sec:problem}

Let an open-world asset $\mathcal{M}$ be represented by localized surface charts $\{c_i\}_{i=1}^N$, where each chart has a spatial support, a local frame, and neighboring candidate contacts. A compression-centric tokenizer would map the asset to a flat sequence of geometry codes and leave component membership or attachment validity implicit. Our goal is different: the generative representation should remain \emph{operational}, answering three decoding-time queries: what local shape a chart contains, which macro-component owns it, and which neighboring charts form plausible physical seams.

C2LT-3D therefore represents an object as a discrete relational graph
\begin{equation}
    \mathcal{G}_{\mathrm{C2LT}} =
    \Big(\{t_i,p_i,h_i,\mathbf{T}_i,s_i\}_{i=1}^N,
    \{e_{ij},\hat{C}_{ij},\Delta\hat{T}_{ij},\hat{y}^{\mathrm{coll}}_{ij}\}_{(i,j)\in\mathcal{E}}\Big).
\end{equation}
Here $t_i=(t_i^{\mathrm{geo}},t_i^{\mathrm{bnd}})$ denotes the local geometry and boundary token streams, $p_i$ is an unsupervised partition hint, $h_i$ is a contextualized token, $\mathbf{T}_i$ and $s_i$ store local pose and scale, and each edge carries seam compatibility $\hat{C}_{ij}$, relative-pose refinement $\Delta\hat{T}_{ij}$, and collision evidence $\hat{y}^{\mathrm{coll}}_{ij}$. We treat $\mathcal{G}_{\mathrm{C2LT}}$ not merely as an encoding of an object, but as the generative state over which reconstruction, repair, intervention, and constrained decoding are performed. It is not a semantic scene graph; it is a low-level geometric-relational state without semantic part labels.

This definition is deliberately testable. The local token $t_i$ should explain chart geometry without absorbing global pose; the partition variable $p_i$ should suppress interference between nearby components; and the edge variables should become useful precisely when local distance is misleading. The intended invariant is that each state variable is both reconstructive and operational: it contributes to decoding, and it remains addressable for intervention when the decoded geometry alone is ambiguous.

% =========================================================================
% 3. C2LT-3D AS AN INTERFACE-CENTRIC GENERATIVE STATE
% =========================================================================
\section{C2LT-3D: State Construction Under Structural Operators}
\label{sec:method}

\begin{figure}[t]
\centering
\includegraphics[width=\textwidth]{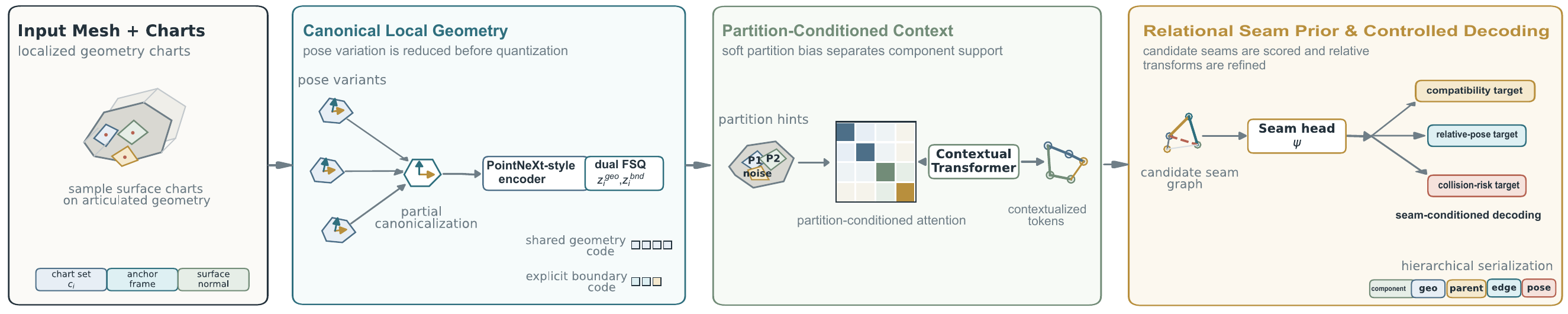}
\caption{\textbf{The C2LT-3D architecture.} The generative state is organized into canonical local geometry, partition-conditioned context, and a relational seam prior for controlled decoding. Local charts are encoded into reusable geometry codes, contextualized by unsupervised partition hints, and scored by a seam head that predicts compatibility, relative-pose refinement, and collision risk for token-space repair and structural intervention.}
\label{fig:method}
\end{figure}

The architecture in Figure~\ref{fig:method} has three coupled but deliberately separated roles. First, a chart-local tokenizer learns reusable geometry codes in canonical frames. Second, a partition-conditioned context model turns independent local codes into object-contextualized tokens while keeping component ownership explicit. Third, a seam prior scores whether two contextualized charts can physically attach and how their relative transform should be refined. The factorization is not introduced for performance alone: each component targets a distinct failure mode that is not natively exposed by spatial compression alone. Local geometry should not be forced to encode global assembly, nearby components should not leak into each other, and attachment validity should not be inferred only from surface similarity.

\subsection{Geometry, Context, and Assembly Factorization}
\label{sec:factorization}

\textbf{Canonical local geometry.}
For each chart $c_i$ with local points $\mathbf{X}_i$, anchor $\mathbf{a}_i$, normal $\mathbf{n}_i$, and scale $s_i$, we define a deterministic local frame with rotation $\mathbf{R}_i\in SO(3)$ and pose $\mathbf{T}_i=(\mathbf{R}_i,\mathbf{a}_i)$, and encode the normalized neighborhood
\begin{equation}
    \tilde{\mathbf{X}}_i = s_i^{-1}\mathbf{R}_i^\top(\mathbf{X}_i-\mathbf{a}_i).
\end{equation}
The local $z$-axis follows the anchor normal; the tangent axis is selected by projecting a global reference into the tangent plane, and covariance-based fallbacks are used for planar or symmetric neighborhoods. In all cases the tangent vector is re-orthogonalized against the normal and completed into a right-handed frame. This \emph{partial} canonicalization removes dominant pose variation without discarding the explicit pose residual needed for assembly. A local encoder and finite-scalar quantizer \citep{fsq} map $\tilde{\mathbf{X}}_i$ to geometry and boundary codes, and a local decoder predicts occupancy and normals at chart-local query points. Thus $t_i$ is trained to represent local shape, while $\mathbf{T}_i$ and $s_i$ keep the information needed to place the chart back into object space.

\textbf{Partition-conditioned context.}
Open-world objects often contain components that touch, intersect slightly, or lie close in Euclidean distance. If all charts attend only by spatial proximity, nearby but distinct components can leak into each other. We therefore estimate unsupervised macro-component hints $\Pi=\{\mathcal{P}_1,\dots,\mathcal{P}_K\}$ from connected components, geometric splitting of oversized components, and absorption of small isolated fragments into nearby supports. These hints are not semantic labels and are not assumed to be perfect; they act as soft structural evidence in the contextual Transformer \citep{vaswani2017attention}:
\begin{equation}
    \alpha_{ij} =
    \frac{
    \exp\left(\mathbf{q}_i^\top \mathbf{k}_j/\sqrt{d} + b_{\mathrm{part}}(i, j) + b_{\mathrm{geom}}(i, j)\right)
    }{
    \sum_{\ell}
    \exp\left(\mathbf{q}_i^\top \mathbf{k}_{\ell}/\sqrt{d} + b_{\mathrm{part}}(i, \ell) + b_{\mathrm{geom}}(i, \ell)\right)
    },
\end{equation}
where $b_{\mathrm{part}}$ distinguishes same-partition from cross-partition pairs and $b_{\mathrm{geom}}$ is a learned bias from normalized relative position and scale ratio. The contextualizer first combines each token feature with partition, position, and scale embeddings, then applies attention with these pair biases; its output is $h_i=\mathrm{Context}(t_i,p_i,\mathbf{a}_i,s_i)$. Thus $h_i$ can borrow object-level context while preserving the distinction between same-component support and neighboring-component interference.

\textbf{Relational seam prior.}
The seam head is the assembly operator of the generative state. Candidate edges are proposed between nearby supports, but geometric proximity alone is insufficient: a nearby part may be a valid contact, a collision, or an unrelated adjacent surface. For each candidate pair, the seam head predicts compatibility $\hat{C}_{ij}$, relative-transform refinement $\Delta \hat{T}_{ij}$, and collision likelihood $\hat{y}_{ij}^{\mathrm{coll}}$:
\begin{equation}
    \hat{C}_{ij}, \Delta \hat{T}_{ij}, \hat{y}_{ij}^{\mathrm{coll}} = \text{SeamHead}_{\psi}\Big(h_i, h_j, \Delta T_{ij}^{\mathrm{coarse}}, s_i / s_j\Big).
\end{equation}
Compatibility targets are derived from support overlap, boundary Chamfer score, normal consistency, and local occupancy consistency. We train this head with compatibility regression, transform refinement, collision supervision, and a margin loss that separates invalid or colliding seams from plausible attachments. Because the seam prior consumes contextualized tokens rather than dense meshes, the same learned operator can be used for attachment ranking, latent repair, and decoding-time constraints.

\subsection{Decoding, Repair, and Object-Level Realization}
\label{sec:decoding_realization}

From this perspective, generation is a sequence of constrained state transitions rather than unconstrained token prediction: each proposed chart is contextualized, attached, validated, and optionally repaired through the seam operator. The state supports reconstruction from known chart tokens, latent repair by ranking and refining candidate parents, and controlled sequence decoding of partitions, local tokens, parent links, edge types, and coarse relative transforms. The seam prior acts as an assembly operator by contributing a structural decoding energy
\begin{equation}
    E(s)= -\log p_{\mathrm{AR}}(s)
    + \lambda \sum_{(i,j) \in \mathcal{E}_s}
    \big[-\log \max(\hat{C}_{ij},\epsilon)\big],
\end{equation}
with a small numerical floor $\epsilon>0$, and selects lower-energy candidates. Both terms are non-negative costs: $-\log p_{\mathrm{AR}}(s)$ is the sequence cost, while the bracketed term is a seam penalty that is near zero for high-compatibility seams and large for invalid seams. Generated structures are therefore guided not only by token likelihood but also by local assembly plausibility. This is still a local relational constraint rather than a proof of global topological validity; global loop closure can require downstream refinement, as discussed in Appendix Section~\ref{sec:supp_hypothesis}.

Object-level reconstruction requires a final deterministic realization map from chart-local fields to an assembled support. A direct union of chart-local occupied points can thicken boundaries and create detached fragments even when each local field is accurate, because independently decoded charts overlap in object space without agreeing on which component owns the boundary. We therefore realize the object through component-owned support projection. Let $\tilde{\mathbf{q}}$ be a decoded support sample in the local coordinate frame of chart $i$, and let $\mathbf{q}=\mathbf{a}_i+s_i\mathbf{R}_i\tilde{\mathbf{q}}\in\mathbb{R}^3$ be the same candidate point after placing it in the final object coordinate system. The realization map keeps $\mathbf{q}$ in the final object-level support for chart $i$ only when its nearest structural support remains owned by the same unsupervised component,
\begin{equation}
    d_{\mathrm{own}}(\mathbf{q}) \le d_{\mathrm{other}}(\mathbf{q}) + m,
\end{equation}
where ``component'' refers to the unsupervised macro-component or partition that owns chart $i$, not to a semantic part label. Thus $d_{\mathrm{own}}$ is the nearest distance from $\mathbf{q}$ to the support of chart $i$'s owning partition, and $d_{\mathrm{other}}$ is the nearest distance to supports owned by all other partitions. The margin $m$ is fixed in the evaluator, and a per-component keep floor protects thin structures. This realization map introduces no learned parameters; it simply uses the component ownership exposed by the representation, so the open-world reconstruction table evaluates the interface-centric generative state rather than a second learned mesher. Appendix Table~\ref{tab:realization_sensitivity} studies the fixed margin and keep-floor choices.

For applications that require an explicit triangle mesh rather than a support surface, we also evaluate an additional mesh-token realization decoder: a small causal decoder maps C2LT chart prefixes to adjacent triangle-strip mesh tokens under teacher forcing. This mesh realization is not used for the open-world reconstruction table; it isolates whether the exposed chart state can condition a faithful triangle detokenizer.

\paragraph{Optimization protocol.}
C2LT-3D is optimized in three phases that mirror the factorization. \textbf{Tokenizer training} learns the encoder, FSQ tokenizer, and local field decoder on isolated canonical charts, establishing reusable local geometry tokens. \textbf{Context training} freezes the tokenizer and trains the contextual Transformer and object realization map on partitioned objects, so local codes become component-conditioned without changing the codebook. \textbf{Seam-prior training} fits the compatibility head and then performs a short compatibility-focused continuation over the seam/context state. The tokenizer, contextualizer, seam prior, and deterministic object realization are therefore evaluated as one interface-centric generative state; the mesh-token realization decoder is presented separately as an application-level evaluation. The inference-time execution diagram appears in Section~\ref{sec:supp_inference_overview} and Figure~\ref{fig:supp_inference_overview}; the core metric contract is summarized in Section~\ref{sec:arxiv_metric_contract}; and full layer specifications, preprocessing choices, optimization schedules, and complete metric definitions are provided in Appendix Sections~\ref{sec:supp_details} and~\ref{sec:supp_metrics}.

% !TEX root = main.tex

\section{Operational State Transitions}
\label{sec:supp_formalization}

To make the interface-centric view concrete before the experiments, we spell out the state variables and transitions used by C2LT-3D.

\textbf{Tokenization map.} Let the continuous manifold $\mathcal{M}$ be sampled into discrete charts $c_i \sim \mathcal{S}(\mathcal{M})$. A learned map $f_\theta: c_i \rightarrow t_i$ embeds this localized geometry into separate geometry and boundary token streams $t_i=(t_i^{\mathrm{geo}},t_i^{\mathrm{bnd}})$. The corresponding context tokens are $h_i = \text{Contextualize}(t_i, p_i, \mathbf{a}_i, s_i)$, where $p_i$ is the unsupervised partition hint and $(\mathbf{a}_i,s_i)$ provide the chart position and scale used by the pair-biased contextualizer.

\textbf{Candidate edge proposal $\mathcal{E}$.} The proposal function identifies structurally viable neighbors before expensive latent compatibility scoring. A candidate edge $e_{ij}$ is proposed if the minimum distance between the spatial supports of $c_i$ and $c_j$ satisfies $D(c_i, c_j) < \epsilon_{\mathrm{contact}}$.

\textbf{Generative state transition.} The partial object state is represented by a token graph $G_t=(V_t, E_t)$. A valid transition $\mathcal{T}(G_t, t_{\mathrm{new}})$ occurs when generating a new token $t_{\mathrm{new}}$ and validating its required incoming edges against the existing graph via the assembly operator $\psi$.

\textbf{Sequence termination.} In the controlled serialization study, the sequence $G=(V,E)$ terminates when an explicit `[EOS]' token is sampled or the fixed context budget is reached. Seam scores are used to rank and refine candidate attachments during decoding; they are not treated as hard guarantees that all possible physical valencies have been exhausted.

\textbf{Local validity vs. global closure.} We distinguish local validity from global closure. The state scores local edge validity through seam compatibility constraints such as surface overlap and normal continuity. However, due to accumulated transform error and structural loops, local validity alone does not guarantee global geometric coherence. We therefore interpret seam constraints as local relational evidence that can be paired with collision checks or downstream refinement, rather than as a complete global-closure proof.

\subsection{Inference-Time Overview}
\label{sec:supp_inference_overview}

\begin{figure}[t]
\centering
\includegraphics[width=\textwidth]{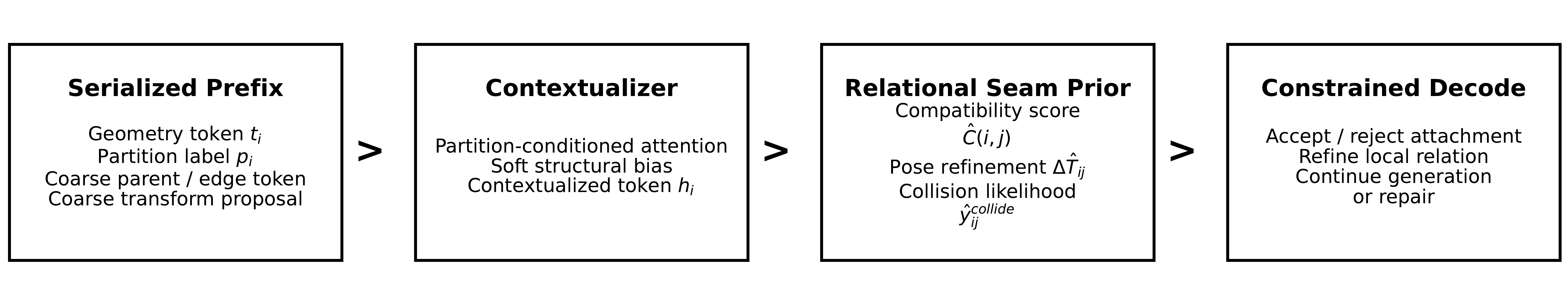}
\caption{\textbf{Inference-time overview of C2LT-3D.} The model first contextualizes chart-local geometry tokens with partition-conditioned attention, then applies the seam prior to score and refine candidate local attachments. The same inference-time state update underlies both controlled decoding and the latent repair experiments.}
\label{fig:supp_inference_overview}
\end{figure}

Figure~\ref{fig:supp_inference_overview} summarizes the inference-time pipeline. A serialized prefix provides geometry tokens, partition labels, and coarse attachment proposals. The contextualizer updates those local codes under partition-conditioned attention. The seam prior then scores and refines candidate seams, and the decoder either accepts the refined attachment or suppresses it during controlled decoding or repair.

\subsection{Evaluation Contract and Core Structural Metrics}
\label{sec:arxiv_metric_contract}

The reconstruction tables use paired fixed-object evaluation: every method is evaluated on the same object IDs, query lattice, object normalization, and threshold convention. The structure-sensitive scores are therefore interpreted as controlled diagnostics rather than stand-alone physical measurements. We always report them with standard geometry measures, object-level visualizations, bootstrap intervals, and robustness slices; Appendix Section~\ref{sec:supp_metrics} gives the full evaluator, edge cases, and additional downstream metric definitions.

\textbf{Component Separation Score.} Let $\hat{X}_k$ be the predicted point set owned by component $k$ and $d(x,A)=\min_{a\in A}\|x-a\|_2$. With the object-adaptive threshold $\tau=\max(0.02\cdot\mathrm{extent},10^{-3})$, the separation score measures whether predicted components remain distinct rather than merging into each other:
\begin{equation}
S_{\mathrm{sep}}=1-\frac{1}{|\mathcal{P}|}\sum_{(i,j)\in\mathcal{P}}
\frac{1}{2}\left[
\frac{1}{|\hat{X}_i|}\sum_{x\in \hat{X}_i}\mathbf{1}\{d(x,\hat{X}_j)<\tau\}
+\frac{1}{|\hat{X}_j|}\sum_{x\in \hat{X}_j}\mathbf{1}\{d(x,\hat{X}_i)<\tau\}
\right],
\end{equation}
where $\mathcal{P}$ is the set of nonempty valid predicted component pairs. Higher values indicate less predicted cross-component overlap.

\textbf{Cross-Component Contamination Rate.} Contamination measures whether predicted points owned by one component leak toward another ground-truth component. Let $X_k$ be the ground-truth support for component $k$ and $X_{\neg k}$ the union of all other supports:
\begin{equation}
C_{\mathrm{rate}}=\frac{\sum_k\sum_{x\in\hat{X}_k}\mathbf{1}\left[
d(x,X_{\neg k})+\tau<d(x,X_k)\;\vee\;\left(d(x,X_{\neg k})<\tau \wedge d(x,X_k)>\tau\right)
\right]}{\sum_k|\hat{X}_k|}.
\end{equation}
Lower values indicate less cross-component leakage. Empty and degenerate cases follow the fixed conventions in Appendix Section~\ref{sec:supp_metrics}, so no method receives a method-specific threshold or exception.

% !TEX root =  neurips_2026.tex

% =========================================================================
% 5. EXPERIMENTS
% =========================================================================
\section{Experiments}
\label{sec:experiments}
\setlength{\textfloatsep}{5pt}
\setlength{\floatsep}{5pt}
\setlength{\intextsep}{5pt}

We evaluate whether factorizing local shape, structural isolation, and assembly relations yields an interface-centric generative state that remains useful beyond surface compression. The study asks whether structure survives zero-shot transfer to open-world assets, whether the same tokens support latent-space relational repair, and whether the state can be serialized and constrained under a fixed lightweight decoder. Our aim is not to replace every mesh generator; it is to test whether the representation exposes operations that compression-centric tokens leave implicit.

\textbf{Evaluation Protocol.} We train C2LT-3D exclusively on ShapeNet \citep{shapenet}, using a 40,000-object training subset drawn from a 44,473-object strictly watertight, single-component train pool, together with 2,535 validation objects and 2,510 held-out test objects; the full preprocessed ShapeNet corpus therefore contains 49,518 objects. For zero-shot evaluation, we use Objaverse \citep{objaverse} through a geometry-clean structural benchmark derived from the same normalized source root, retaining 42,100 assets out of 43,226 candidates (97.4\% kept). The fixed-object evaluation measures object-level structural reconstruction: each method encodes the same source asset, realizes a source-frame surface or mesh representation, and is scored against the source geometry with the same query lattice and structural evaluator. For auditability, Appendix Sections~\ref{sec:supp_bpt_subset}, \ref{sec:supp_metrics}, \ref{sec:supp_details}, and~\ref{sec:supp_responsible_use} document the split, metrics, implementation details, and release scope.

\textbf{Baseline Roles.} BPT and VQ-Patch are same-protocol performance baselines on the identical object IDs and evaluator: BPT is external compression-centric, and VQ-Patch removes partition conditioning and seam modeling. MeshAnythingV2, MeshGPT, and LoST are released-interface controls (Appendix Tables~\ref{tab:baseline_scope} and~\ref{tab:external_sota_audit}); they test whether released outputs natively expose ownership, attachment validation, and repairable seams rather than replacing the fixed-object performance baselines.

\begin{figure}[t]
\centering
\includegraphics[width=0.96\textwidth]{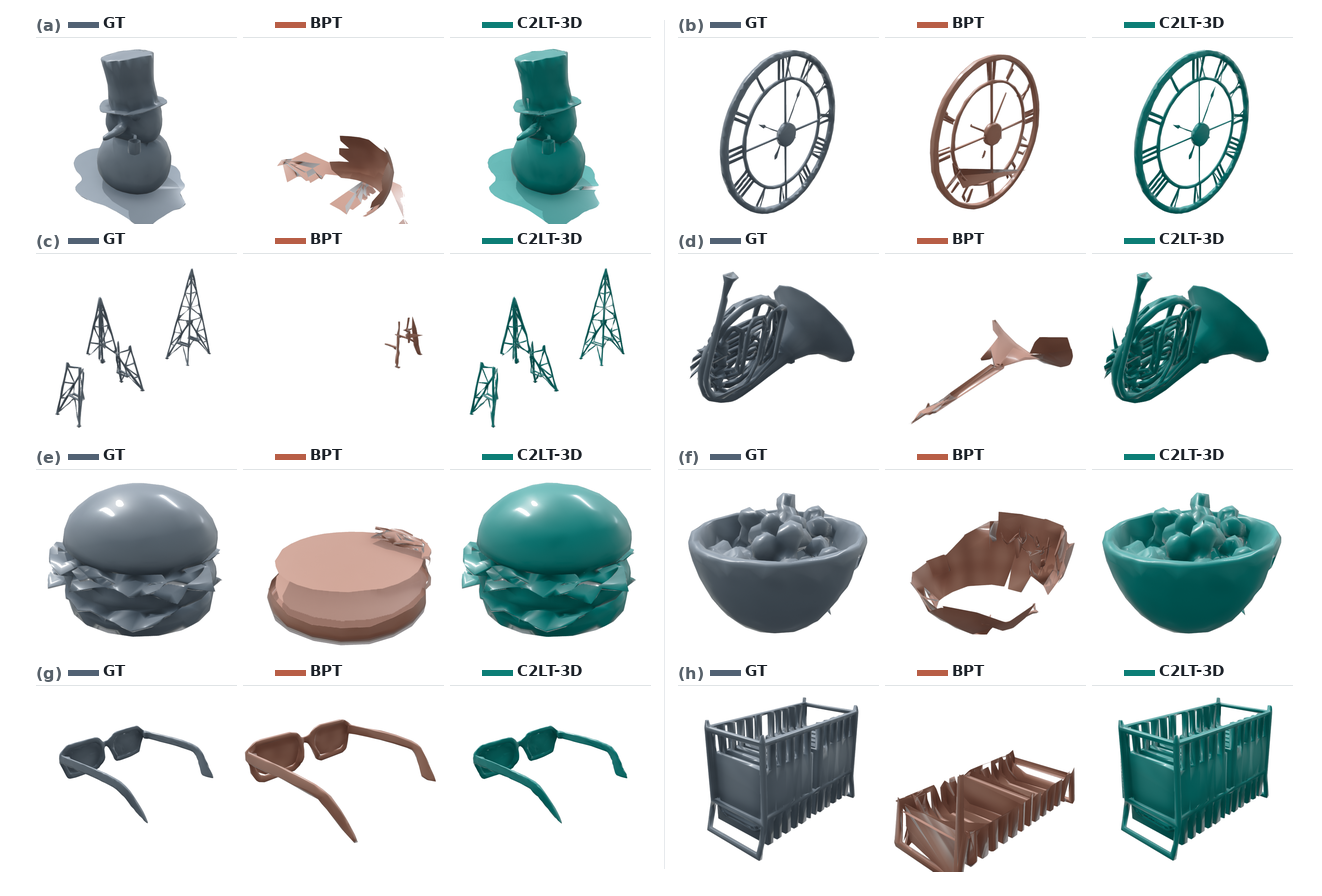}
\caption{\textbf{Object-level GT--BPT--C2LT-3D mesh comparison on open-world assets.} Eight fixed Objaverse-LVIS objects cover diverse structures and representative BPT artifacts, including detached and merged geometry. C2LT-3D is rendered through the explicit mesh-token realization decoder conditioned on C2LT chart states; Table~\ref{tab:reconstruction} remains measured with deterministic component-owned realization. Source-up orientation is preserved, and no post-decoding mesh repair is applied to C2LT panels.}
\label{fig:openworld_object_qualitative}
\end{figure}

\textbf{Evaluation Design.} We keep three evidence paths deliberately separate. Quantitative reconstruction uses the deterministic component-owned realization map, so Table~\ref{tab:reconstruction} measures the interface-centric generative state without adding a learned mesher. Latent repair is evaluated directly in token space, so Table~\ref{tab:repair_main} tests attachment reasoning rather than surface resampling quality. Mesh-token realization appears in Figure~\ref{fig:openworld_object_qualitative}, Section~\ref{sec:supp_explicit_mesh_realization}, and Table~\ref{tab:explicit_mesh_realization_full}; it is a state-to-mesh realization study, not a claim of state-of-the-art unconditional mesh generation.

\subsection{A Benchmark for Open-world Structural Robustness}
\label{sec:exp_robustness}

Table~\ref{tab:reconstruction} evaluates whether an interface-centric generative state transfers more robustly than compression-centric tokenization to unseen open-world multi-component assets. The task is fixed-object structural reconstruction from encoded 3D assets, not point-cloud completion: every method receives the same source object, and the realized surface or mesh is compared to the source geometry in the same reference frame. We use a geometry-clean Objaverse-LVIS split that emphasizes intersecting multi-component geometry and report 1,024 unseen assets sampled once from this split under the same protocol for every method; coarse full-split rows and detailed runtime breakdowns are provided in the appendix.
BPT serves as a released high-capacity compression-centric mesh-token baseline under this open-world evaluator. We additionally include an in-house \emph{VQ-Patch (Spatial-Only)} baseline, implemented as a two-stream spatial patch tokenizer in the spirit of VQ-VAE/VQ-GAN style discrete autoencoding \citep{vqvae,vqgan}. It shares the local geometry and boundary code streams with C2LT-3D, but removes partition conditioning and seam modeling, giving a stronger pure spatial-compression reference on the same 1,024 assets; Appendix Table~\ref{tab:vqpatch_spec} specifies its architecture. Appendix Table~\ref{tab:external_sota_audit} complements these rows with a released-interface comparison: MeshAnythingV2 receives privileged GT-derived point-normal inputs, is mapped back to the source frame, and is scored by the same evaluator. It can produce plausible meshes, but its released output contains no component ownership, attachment variables, or queryable seam-repair operator. For C2LT-3D, object-level reconstruction uses a deterministic component-owned realization map that retains predicted query points assigned to their own component support, with a 90\% per-part keep floor; it uses no learned parameters or retraining.

To move beyond generic geometry distances, we separate structural errors into cross-component leakage, boundary erosion, spurious seam prediction, collision-blind attachment, and support violation. We then quantify these errors with structure-sensitive measures, most notably \emph{Component Separation Score} and \emph{Cross-Component Contamination Rate} (Section~\ref{sec:arxiv_metric_contract}; full evaluator in Appendix Section~\ref{sec:supp_metrics}), so that Table~\ref{tab:reconstruction} measures not only reconstruction fidelity but structural robustness under open-world transfer. To avoid over-interpreting any custom score, we place these measures alongside standard geometry metrics, bootstraps, and object-level mesh views; the appendix gives the complete formulas, threshold convention, interpretation scope, and a visual witness pairing metric families with visible failure modes.

\begin{table}[t]
\centering
\caption{\textbf{Zero-Shot Open-World Structural Robustness.} Evaluated on 1,024 unseen multi-component assets sampled from the geometry-clean Objaverse-LVIS split. All methods are measured on the same assets under the same structural evaluator, and decode time is shown per object. Boldface highlights the best quality scores and fastest decode time.}
\label{tab:reconstruction}
\resizebox{\textwidth}{!}{
\begin{tabular}{l|cccccc}
\toprule
 & \multicolumn{6}{c}{\textbf{Objaverse-LVIS}} \\
\cmidrule(lr){2-7}
\textbf{Method} & \textbf{Chamfer $\downarrow$} & \textbf{Hausdorff $\downarrow$} & \textbf{Contamination $\downarrow$} & \textbf{Separation $\uparrow$} & \textbf{Norm. Cons. $\uparrow$} & \textbf{Decode / obj $\downarrow$} \\
\midrule
BPT (Spatial Comp.) \citep{bpt} & 0.0614 & 0.4356 & 0.1161 & 0.9220 & 0.0569 & 87.42s \\
VQ-Patch (Spatial-Only) & 0.0303 & 0.3277 & 0.0727 & 0.9608 & 0.3443 & \textbf{0.0009s} \\
\textbf{C2LT-3D (Ours)} & \textbf{0.0268} & \textbf{0.2282} & \textbf{0.0141} & \textbf{0.9780} & \textbf{0.4638} & 0.0332s \\
\bottomrule
\end{tabular}
}
\end{table}

The spatial-only baseline clarifies the mechanism behind the gains. Relative to BPT, VQ-Patch substantially improves geometry and normal consistency, showing that stronger pure compression closes part of the gap. C2LT-3D still improves every quality metric over VQ-Patch: Chamfer drops from 0.0303 to 0.0268, Hausdorff from 0.3277 to 0.2282, contamination from 0.0727 to 0.0141, separation rises from 0.9608 to 0.9780, and normal consistency from 0.3443 to 0.4638. VQ-Patch is fastest because it uses the lightest spatial realization path, but this speed does not translate into stronger structural fidelity; BPT is far more expensive and lower-scoring, whereas C2LT-3D decodes the same 1,024 objects in 34.05 seconds total ($\sim$0.0332s/object). Appendix Tables~\ref{tab:same_filter_control}, \ref{tab:complexity_slice}, and~\ref{tab:filtered_full5000} provide a factorized realization fairness analysis, a high-complexity slice, and a 5,000-object coarse benchmark. The seam prior is relational, so we evaluate it below through attachment ranking rather than local reconstruction.

\begin{table}[t]
\centering
\caption{\textbf{Capability gap beyond surface compression.} BPT and VQ-Patch use the fixed-object reconstruction protocol in Table~\ref{tab:reconstruction}; MeshAnythingV2 is a privileged geometry-only comparison in Appendix Figure~\ref{fig:external_meshanythingv2_audit}: it receives GT-derived point-normal input, yet returns mesh geometry rather than ownership, attachment variables, or repairable seam state.}
\label{tab:capability_gap}
\scriptsize
\setlength{\tabcolsep}{2.6pt}
\renewcommand{\arraystretch}{1.05}
\resizebox{\textwidth}{!}{
\begin{tabular}{P{0.20\textwidth}|P{0.18\textwidth}P{0.18\textwidth}P{0.19\textwidth}P{0.20\textwidth}}
\toprule
\textbf{Native capability} & \textbf{BPT} & \textbf{VQ-Patch} & \textbf{MeshAnythingV2} & \textbf{C2LT-3D} \\
\midrule
Expose component ownership & Not exposed by compressed geometry tokens. & Not exposed; spatial-only code stream. & Not exposed in released mesh output. & Native partition state $p_i$ and component-owned realization. \\
Validate attachments during decoding & Not exposed natively. & Not exposed natively. & Not exposed natively; mesh-only output. & Native seam prior over candidate chart relations. \\
Repair invalid latent seams & No latent seam operator. & No latent seam operator. & No latent state for seam repair. & Hard Valid@1 $0.6669$; Heur.-Fail Valid@1 $0.6675$. \\
Preserve reference-frame support & Contam.\ $0.1161$; Sep.\ $0.9220$. & Contam.\ $0.0727$; Sep.\ $0.9608$. & Source-frame Contam.\ $0.1173$; Sep.\ $0.9292$. & Contam.\ $0.0141$; Sep.\ $0.9780$. \\
\bottomrule
\end{tabular}
}
\end{table}

To assess whether the comparison is stable beyond aggregate means, the appendix also includes a paired fixed-object bootstrap over the same 1,024 assets. All five C2LT-3D quality improvements over VQ-Patch and BPT have strictly positive 95\% confidence intervals; against VQ-Patch, the per-object win rates are 84.8\% for Chamfer, 88.0\% for Hausdorff, 99.5\% for contamination, 90.9\% for separation, and 96.9\% for normal consistency (Appendix Table~\ref{tab:matched_bootstrap}).

\subsection{Latent Structural Repair and Intervention}
\label{sec:exp_assembly}

If Table~\ref{tab:reconstruction} evaluates transfer robustness, this section asks whether the generative state can identify and repair invalid attachment structure before reconstructing or remeshing the object. We therefore use the seam prior as a latent assembly operator: it ranks candidate parents and refines relative transforms from stored C2LT state variables rather than from a newly decoded surface. Latent repair is not an auxiliary downstream task; it is a diagnostic for whether attachment-relevant variables remain exposed inside the representation rather than entangled in reconstructed geometry. Success on this task is less consistent with a compression-only explanation: it requires the representation to expose variables that can be directly acted on, not only local geometry that can be reconstructed. For this reason, we treat latent repair and structural intervention as the most direct evidence that C2LT-3D is an interface-centric generative state, not merely a better local decoder.

We make this bridge explicit with a latent repair benchmark derived from the same geometry-clean Objaverse-LVIS source split used in Table~\ref{tab:reconstruction}. Rather than sampling a small number of prefix corruptions, we build an inter-part edge bank: every valid cross-component attachment edge becomes a detached-child repair task, and the model must rank candidate parents using only the stored C2LT chart, partition, and seam variables. This yields 1,964 repair tasks across 164 objects, including 1,360 hard tasks where local geometry scores an invalid parent above the valid target and 1,194 heuristic-fail tasks where the nearest-neighbor top-1 parent is invalid. Table~\ref{tab:repair_main} shows the capability gap. On the hard subset, the learned seam prior raises Valid@1 from 0.1221 for geometric nearest-neighbor ranking and 0.2456 for a dense support verifier to 0.6669. The same pattern holds on the stricter heuristic-fail subset, where the geometric nearest-neighbor baseline has Valid@1 equal to zero by construction and C2LT-3D reaches 0.6675. Appendix Table~\ref{tab:repair_rank_ci} gives bootstrap confidence intervals, Top-$k$ recall, and MRR. These results provide direct quantitative evidence that the seam prior functions as an intervention operator over attachment structure rather than only as a passive compatibility regressor.

The absolute hard-repair numbers should be read as an adversarial-regime result, not as a saturated repair benchmark. On the full edge bank, local nearest-neighbor geometry remains competitive because many attachments are spatially obvious; the hard and heuristic-fail subsets focus on the non-obvious cases.

\begin{table}[H]
\centering
\caption{\textbf{Capability gap in latent structural repair.} \emph{All} contains 1,964 inter-part edge-target repair tasks; \emph{Hard} contains the 1,360 tasks where local geometry ranks an invalid parent above the valid target; \emph{Heur.-Fail} contains the 1,194 tasks where the nearest-neighbor top-1 parent is invalid. Dense support is an external verifier, not a native generative state.}
\label{tab:repair_main}
\setlength{\tabcolsep}{4pt}
\footnotesize
\begin{tabular}{l|cccc}
\toprule
\textbf{Method} & \textbf{All Valid@1 $\uparrow$} & \textbf{Hard Valid@1 $\uparrow$} & \textbf{Hard Parent@1 $\uparrow$} & \textbf{Heur.-Fail Valid@1 $\uparrow$} \\
\midrule
Geometric NN & 0.3921 & 0.1221 & 0.0000 & 0.0000 \\
Dense Support Verifier & 0.2337 & 0.2456 & 0.0971 & 0.2420 \\
C2LT-3D Seam Head Only & 0.2587 & 0.2397 & 0.1103 & 0.2437 \\
C2LT-3D w/o Partition & 0.1309 & 0.0765 & 0.0544 & 0.0645 \\
\textbf{C2LT-3D Learned Seam Prior} & \textbf{0.6202} & \textbf{0.6669} & \textbf{0.2713} & \textbf{0.6675} \\
\bottomrule
\end{tabular}
\end{table}

Dense support verification is intentionally local and non-generative: it can reject some invalid attachments, but it does not provide a reusable latent state or learned assembly prior, which explains its limited full-pool performance. Because several candidate parents may form geometrically valid seams for the same child, Valid@1 is the primary repair metric, while Parent@1 measures recovery of the particular held-out edge target.

\begin{figure}[t]
\centering
\includegraphics[width=0.91\textwidth]{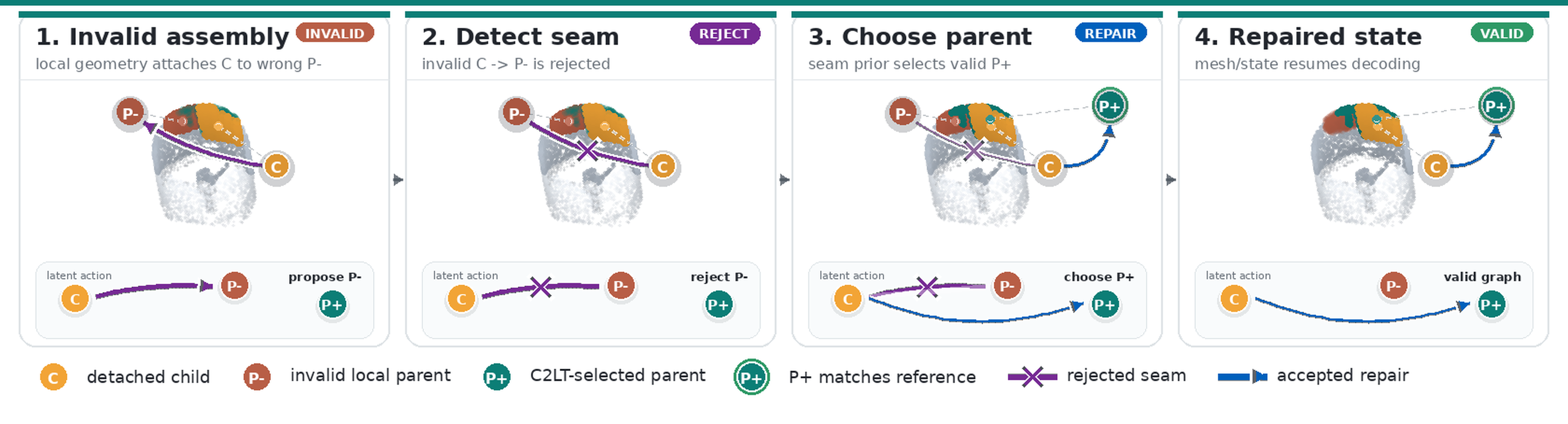}
\caption{\textbf{Object-level latent repair trace.} Left to right: the upper view highlights the proposed, rejected, repaired, or accepted seam, while the lower inset shows the latent graph operation. C2LT-3D rejects \textbf{C}$\rightarrow$\textbf{P-} and selects \textbf{P+}; the green halo marks the held-out valid reference.}
\label{fig:repair}
\end{figure}

Figure~\ref{fig:repair} visualizes the repair decision as a state transition. Starting from an invalid local attachment, the exposed seam state lets C2LT-3D reject the wrong parent, select a valid parent, and continue decoding from a repaired structural state.

Together, Tables~\ref{tab:reconstruction}--\ref{tab:repair_main} separate two notions that compression-centric tokenizers conflate: reconstructing local geometry versus preserving attachment-relevant structure. Table~\ref{tab:reconstruction} measures whether structure survives transfer, while Table~\ref{tab:repair_main} measures whether that structure remains usable for relational reasoning when local geometry is misleading or adversarial.

\subsection{Isolating Structural Mechanisms (Causal Ablation)}
\label{sec:exp_causal_ablation}

To test whether the gains arise from factorization rather than merely from model capacity, we remove one module at a time and ask whether each ablation selectively damages the metric family it is designed to control. \textbf{Canonicalization} primarily affects tokenizer-local state: without it, geometry-token perplexity drops from 5388.1 to 3369.9 and boundary-token perplexity from 728.8 to 397.2, while occupancy BCE worsens from 0.259 to 0.330, patch IoU falls from 0.657 to 0.583, and masked normal cosine decreases from 0.582 to 0.106 (Appendix Table~\ref{tab:canonicalization_ablation}). \textbf{Partition-conditioned context} controls structural isolation: removing it selectively increases \emph{Cross-Component Leakage}, degrades distance and normal-consistency metrics, and lowers hard-repair accuracy. \textbf{The relational seam prior} is exposed most clearly by assembly-sensitive metrics: on the hard repair subset in Table~\ref{tab:repair_main}, the learned seam prior raises Valid@1 from 0.2397 for the seam-head-only variant and 0.0765 without partition conditioning to 0.6669; exact Parent@1 rises from 0.1103 to 0.2713. The resulting error pattern is dissociated rather than uniform: canonicalization improves shape--pose separation, partition conditioning suppresses cross-component leakage, and the seam prior improves attachment ranking and repair. This selectivity is less consistent with a capacity-only explanation.

\subsection{Controlled Serialization Study}
\label{sec:exp_generation}

Finally, we ask whether the same representation remains usable as a \emph{serialized structural state}. This evaluation is intentionally narrower than our robustness and repair studies: it isolates representation quality under a fixed 6-layer, 256-dimensional decoder-only transformer \citep{vaswani2017attention} rather than claiming the strongest end-to-end mesh generator. Appendix Table~\ref{tab:generation} shows that the full C2LT-3D state improves structure-sensitive FID \citep{heusel2017gans} from 0.3920 for BPT tokens and 0.3890 without partition conditioning to 0.2893, and gives the best component-intersection quality (0.3074). Boundary clarity is highest for the no-partition variant, but the full state still improves over BPT tokens (0.6020 vs.\ 0.5024). This supports C2LT-3D as a compact generative state whose variables can be serialized, scored, and constrained. The serialization study is separate from the mesh-token realization decoder in Section~\ref{sec:supp_explicit_mesh_realization}, which converts known C2LT chart states into explicit triangles with 99.98\% teacher-forced token accuracy.

Together, Table~\ref{tab:reconstruction}, Table~\ref{tab:repair_main}, and the serialization study test transfer, repair under adversarial local geometry, and compact controlled decoding. All three exercise the same ownership and seam variables under fixed object identities, candidate seams, and evaluator, distinguishing C2LT-3D from a post-hoc reconstruction filter or a purely spatial compression code.

% !TEX root = main.tex

\section{Explicit Mesh-Token Realization Decoder}
\label{sec:supp_explicit_mesh_realization}

\begin{table}[t]
\centering
\caption{\textbf{Explicit mesh-token realization decoder.} The decoder is conditioned on C2LT chart prefixes and predicts adjacent triangle-strip mesh tokens. The model is selected by teacher-forced token accuracy; object-level metrics are computed on the eighteen-object gallery in Appendix Figure~\ref{fig:explicit_mesh_realization}, which is non-overlapping with the main object-level figure.}
\label{tab:explicit_mesh_realization_full}
\scriptsize
\setlength{\tabcolsep}{3pt}
\renewcommand{\arraystretch}{0.66}
\resizebox{0.98\textwidth}{!}{
\begin{tabular}{P{0.25\textwidth}|P{0.29\textwidth}|P{0.39\textwidth}}
\toprule
\textbf{Aspect} & \textbf{Setting / value} & \textbf{Interpretation} \\
\midrule
Mesh-token representation & Adjacent triangle-strip tokens in the normalized C2LT frame; 256 coordinate bins; slot embeddings distinguish first-triangle coordinates, continuation coordinates, strip switches, and EOS. & Provides a mesh-native sequence analogous to contemporary autoregressive mesh tokenizers, but conditioned by the C2LT generative state. \\
Decoder & 8-layer causal transformer \citep{vaswani2017attention}, width 512, 8 heads, context length 1024, sliding stride 256, dropout 0.0, C2LT-prefix conditioning, 28.52M parameters. & A compact mesh realization model; the structural state is supplied by C2LT rather than inferred from raw mesh geometry alone. \\
Validation-selected decoder & Selected at 6,000 optimization steps; cross-entropy 0.001422, perplexity 1.00142, teacher-forced token accuracy 99.982\%. & Shows that known C2LT chart states nearly determine the mesh-token sequence. \\
Object-level evaluation & Eighteen Objaverse-LVIS objects, mean 62.28 decoded charts/object, all decoded chart meshes valid. Weighted teacher-forced accuracy is 99.950\%. & Tests whether chart-level token accuracy survives object-level merging rather than only isolated patch rendering. \\
Mesh fidelity to chart-token union & Predicted-vs.-GT-chart-union Chamfer 0.0101, p95 0.0190; source-vs.-GT-chart-union Chamfer 0.0103, p95 0.0190. & The predicted merged meshes match the chart-token union at approximately the same scale as the chart-token union matches the source mesh on these examples. \\
Comparison to BPT on the same objects & BPT mean Chamfer is 0.2289 and mean Hausdorff is 0.8129 under the Table~\ref{tab:reconstruction} evaluator; the explicit mesh-token realization has mean source Chamfer 0.0103 on the same eighteen-object set. & This is not a replacement for the main benchmark, but it verifies that the mesh-native decoder can produce visually faithful triangle meshes on diverse structures where BPT meshes often fragment or lose parts. \\
Post-processing policy & No post-decoding mesh repair, spike filtering, or topology cleanup is applied in the figure. & Avoids hiding decoder errors; earlier cleanup can remove valid faces on thin structures, so the figure shows the direct mesh-token decoding output. \\
\bottomrule
\end{tabular}
}
\end{table}

The main reconstruction table uses deterministic component-owned object realization to evaluate the interface-centric state without adding a second learned decoder. We separately test whether the exposed chart-local state can condition an explicit triangle-mesh detokenizer. Each chart mesh is serialized as C2LT-frame adjacent triangle-strip tokens; a causal transformer \citep{vaswani2017attention} receives the C2LT chart prefix, predicts the sequence under teacher forcing, and merges decoded chart meshes at object level. This downstream mesh-realization evaluation is not used for Table~\ref{tab:reconstruction}; the additional eighteen-object gallery in Appendix Figure~\ref{fig:explicit_mesh_realization} uses open-world objects that do not overlap with the main object-level figure.

\begin{figure}[t]
\centering
\includegraphics[width=\textwidth]{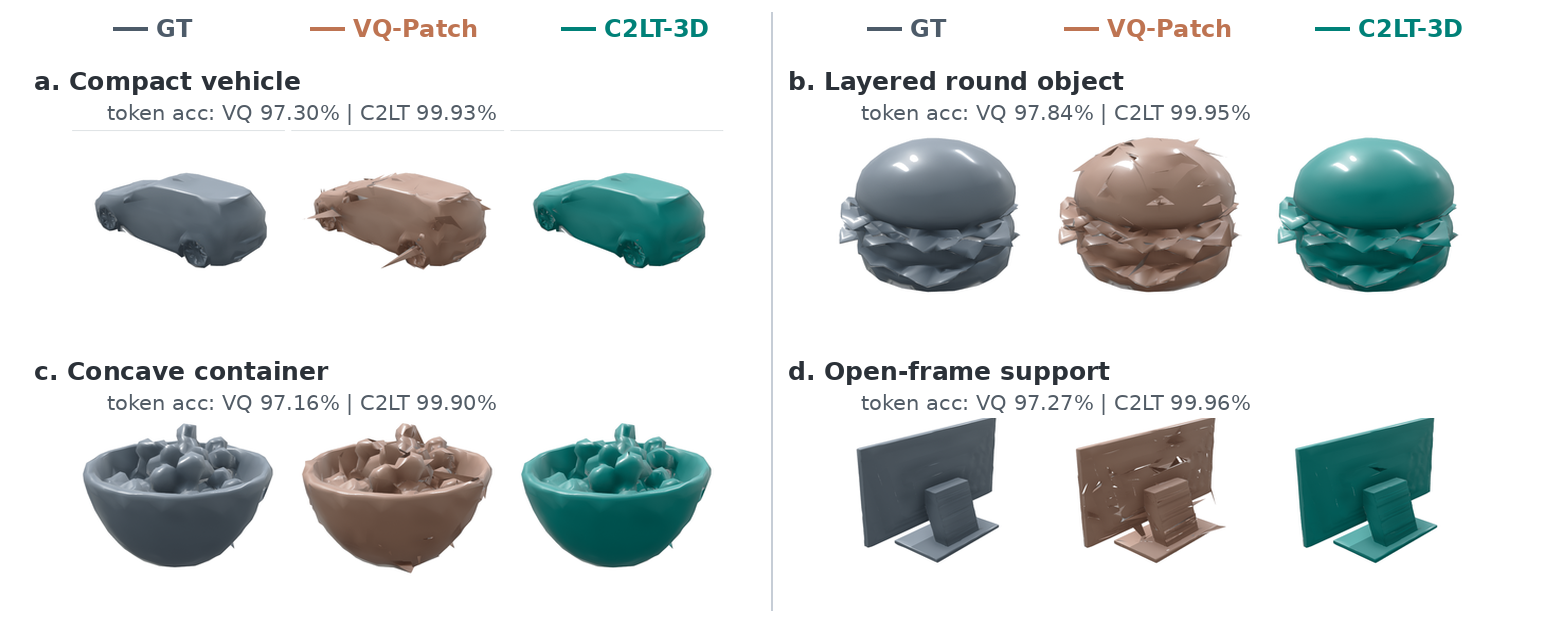}
\caption{\textbf{Conditioning ablation for mesh-token realization.} The adjacent-strip mesh-token decoder, architecture, teacher-forced protocol, and rendering style are fixed; only the conditioning signal changes. Spatial VQ-Patch codes often yield local holes, spikes, or broken sheets despite high token accuracy, whereas C2LT-3D chart states produce more coherent triangle meshes on the same recognizable Objaverse-LVIS objects. No post-decoding mesh repair is applied.}
\label{fig:mesh_realization_vqpatch_ablation}
\end{figure}

To separate the decoder from the representation that conditions it, we also train the same adjacent-strip mesh-token decoder with the spatial-only VQ-Patch chart context. This VQ-Patch-conditioned model reaches 96.98\% validation teacher-forced token accuracy, compared with 99.98\% for the C2LT-conditioned model. On the four-object visualization subset in Figure~\ref{fig:mesh_realization_vqpatch_ablation}, the same decoder gives 97.39\% weighted token accuracy and mean Chamfer 0.0168 under VQ-Patch conditioning, versus 99.93\% and 0.0153 under C2LT conditioning. The visual difference is larger than the Chamfer gap suggests: spatial-only conditioning produces holes, spikes, and broken sheets, while C2LT chart states remain close to the source structure without mesh repair. Table~\ref{tab:explicit_mesh_realization_full} gives the full decoder and evaluation specification.

\FloatBarrier

% !TEX root =  neurips_2026.tex

% =========================================================================
% 6. CONCLUSION
% =========================================================================
\section{Conclusion and Scope}
\label{sec:conclusion}

C2LT-3D instantiates interface-centric generative states for open-world 3D: ownership, attachment validity, and seam repair are explicit variables rather than post-hoc geometry. Under a fixed protocol, these variables improve structural transfer and latent repair within offline-preprocessed geometry and without formal global-topology guarantees. The deterministic realization map is a non-learned readout of ownership already present in the state, while the mesh-token decoder is a state-to-mesh capacity test rather than an unconditional generation claim. The central result is representational: for structured 3D generation, a token state should be judged not only by compression fidelity, but by whether it supports validation, repair, and object-level realization in the same reference frame. This makes tokenization an interface design problem.

% %%%%%%%%%%%%%%%%%%%%%%%%%%%%%%%%%%%%%%%%%%%%%%%%%%%%%%%%%%%%
\bibliographystyle{plainnat}
\bibliography{sample}

% %%%%%%%%%%%%%%%%%%%%%%%%%%%%%%%%%%%%%%%%%%%%%%%%%%%%%%%%%%%%
% \appendix
\appendix
% !TEX root =  neurips_2026.tex

\section*{Appendix}

This appendix complements the paper by making the evaluation contract, scope boundaries, and architectural details explicit. It follows the same evidence paths as the main text: open-world structural transfer, operational capability gaps, latent repair, deterministic object realization, and state-to-mesh realization.

\paragraph{Reader guide.}
For a guided reading path, start with Table~\ref{tab:evidence_map}. The generative-state formalization, inference-time overview, core metric contract, and explicit mesh-token realization are now included in the main text. For baseline scope, read Section~\ref{sec:supp_baseline_scope}. For additional state-to-mesh examples, read Section~\ref{sec:supp_mesh_gallery}. For complete metric definitions and robustness checks, read Sections~\ref{sec:supp_metrics}--\ref{sec:supp_bpt_subset}. For mechanism and repair, read Section~\ref{sec:supp_seam_prior}. For reproducibility and architecture, read Section~\ref{sec:supp_details}. Unless stated otherwise, upward arrows indicate larger-is-better metrics, downward arrows indicate smaller-is-better metrics, and all fixed-object tables use fixed object IDs and the same evaluator.

\paragraph{Reproducibility and release scope.}
The supplementary package contains an overview video and a non-code evaluation audit package, rather than executable code, raw data, or checkpoints. To make the evaluation contract inspectable within the paper, the appendix documents the evaluator design, fixed 1,024-object protocol, repair-task construction, result summaries, and released-interface comparison scope. We maintain the corresponding implementation, evaluator scripts, fixed identifiers, repair-task banks, selected summaries, and permitted checkpoints for release after de-anonymization. Raw ShapeNet, Objaverse, Objaverse-LVIS meshes, external baseline weights, and third-party code are not redistributed; full reproduction requires obtaining those assets from their original sources and preserving their license metadata.

\section{Evidence Roadmap}
\label{sec:supp_evidence_map}

Table~\ref{tab:evidence_map} maps each main statement to its measurement and control. The intent is to make the evaluation logic explicit: no single metric is treated as sufficient evidence for the generative-state interpretation.

\begin{table}[!tbp]
\centering
\caption{\textbf{Evidence roadmap.} We summarize where each core statement is evaluated, what the main control is, and how the result should be interpreted.}
\label{tab:evidence_map}
\footnotesize
\renewcommand{\arraystretch}{1.12}
\resizebox{0.98\textwidth}{!}{
\begin{tabular}{P{0.22\textwidth}|P{0.33\textwidth}|P{0.39\textwidth}}
\toprule
\textbf{Claim} & \textbf{Primary evidence} & \textbf{Control and interpretation} \\
\midrule
Structure, not only surface fidelity, should transfer to open-world objects. & Main Table~\ref{tab:reconstruction}; Appendix Tables~\ref{tab:bpt_subset}, \ref{tab:matched_bootstrap}, \ref{tab:same_filter_control}, and \ref{tab:complexity_slice}; Figure~\ref{fig:structural_metric_witness}. & Same 1,024 Objaverse-LVIS IDs, same evaluator, BPT external baseline, VQ-Patch spatial-only and same-realization comparison, paired bootstrap, and visual metric witness. \\
The gap is operational, not only metric-level. & Main Table~\ref{tab:capability_gap}; Appendix Figure~\ref{fig:external_meshanythingv2_audit}; Table~\ref{tab:external_meshanythingv2_diagnostic}. & Mesh outputs may be plausible while still lacking ownership, attachment validation, and repairable seam variables; the MeshAnythingV2 analysis isolates this distinction under privileged point-normal conditioning. \\
The state exposes actionable assembly information. & Main Table~\ref{tab:repair_main}; Appendix Tables~\ref{tab:repair_rank_ci}, \ref{tab:repair_metrics}, and \ref{tab:intervention_stress}. & A 1,964-task edge bank, hard subset, and heuristic-fail subset make local geometry misleading, so repair cannot be explained by nearest-neighbor support alone. \\
The gains are factorized, not a uniform capacity effect. & Main causal ablation section; Appendix Tables~\ref{tab:canonicalization_ablation} and \ref{tab:seam_metrics}. & Canonicalization affects local fields, partition conditioning affects leakage, and the seam prior affects compatibility ranking and repair. \\
Object realization is deterministic. & Main Table~\ref{tab:reconstruction}; Appendix Table~\ref{tab:realization_sensitivity}. & The component-owned realization map has no learned parameters; the mesh-token decoder is evaluated separately as a teacher-forced state-to-mesh capacity test. \\
The state can condition explicit triangle meshes. & Main Figure~\ref{fig:openworld_object_qualitative}; Section~\ref{sec:supp_explicit_mesh_realization}, Table~\ref{tab:explicit_mesh_realization_full}, Figure~\ref{fig:mesh_realization_vqpatch_ablation}, and Appendix Figure~\ref{fig:explicit_mesh_realization}. & Known C2LT chart states are decoded into adjacent-strip triangle meshes without post-decoding repair and compared against the same decoder conditioned on spatial VQ-Patch codes. \\
Controlled serialization is a state-utility study, not an SOTA generation claim. & Appendix Table~\ref{tab:generation}; serialization protocol below. & All states use the same small decoder-only backbone, so the comparison evaluates state utility rather than model scale. \\
\bottomrule
\end{tabular}
}
\end{table}

\section{External Baselines and Released-Interface Comparisons}
\label{sec:supp_baseline_scope}

Table~\ref{tab:baseline_scope} separates two roles for external methods. Fixed-object performance baselines test open-world structural robustness under identical object IDs and metrics. Released-interface comparisons ask a different question: whether a mesh-generation interface, even when it produces plausible geometry, exposes the operational variables required for ownership queries, attachment validation, and latent seam repair.

\begin{table}[!tbp]
\centering
\caption{\textbf{Baseline scope and released-interface comparisons.} Fixed-object performance baselines are scored in the main reconstruction table. Other released systems are analyzed for whether mesh-generation outputs natively expose the structural state operations studied by C2LT-3D.}
\label{tab:baseline_scope}
\footnotesize
\setlength{\tabcolsep}{3pt}
\renewcommand{\arraystretch}{1.12}
\resizebox{\textwidth}{!}{
\begin{tabular}{l|P{0.26\textwidth}|P{0.46\textwidth}}
\toprule
\textbf{Method family} & \textbf{Role in this paper} & \textbf{Evaluation scope} \\
\midrule
BPT \citep{bpt} & Main external compression-centric baseline & Released pretrained mesh-token pipeline can be run on the fixed 1,024 Objaverse-LVIS assets and evaluated under the same structural protocol. Its compressed output does not expose ownership or seam-repair state. \\
VQ-Patch & Spatial-only reference baseline & In-house VQ spatial tokenizer trained and evaluated under the same data and evaluator protocol; isolates pure spatial compression from structural conditioning. \\
MeshGPT / PolyGen \citep{meshgpt,polygen} & Mesh-token generation references & Important closed-world generators for mesh sequences; their released interfaces do not provide a same-object open-world ownership/seam state for the operations evaluated here. \\
MeshAnything / MeshArt \citep{meshanything,meshart} & Released mesh-generation comparisons & Related systems can produce plausible mesh geometry under their native conditioning, but the released outputs are mesh surfaces rather than explicit ownership, attachment, and repair states. \\
LoST \citep{lost} & Related tokenization perspective & Conceptually relevant latent-token perspective; we use it to position the representation question rather than as an operational structural-state control. \\
\bottomrule
\end{tabular}
}
\end{table}

\begin{table}[!tbp]
\centering
\caption{\textbf{External released-system capability-gap evidence.} The comparison asks whether a released system provides the operational state studied here, not merely whether it can output a plausible mesh. MeshAnythingV2 is evaluated under a favorable GT point-normal conditioning setup and mapped back to the source frame; the result is still a geometry-only output without native ownership, attachment validation, or repairable seam variables.}
\label{tab:external_sota_audit}
\scriptsize
\setlength{\tabcolsep}{3pt}
\renewcommand{\arraystretch}{1.12}
\resizebox{\textwidth}{!}{
\begin{tabular}{l|P{0.22\textwidth}|P{0.26\textwidth}|P{0.34\textwidth}}
\toprule
\textbf{System} & \textbf{External control} & \textbf{Native state exposed?} & \textbf{Evidence in this paper} \\
\midrule
BPT \citep{bpt} & Released high-capacity mesh-token baseline evaluated on the fixed 1,024 Objaverse-LVIS objects. & Mesh-token geometry output; no native component ownership or seam-repair operator. & Quantitative external baseline in Table~\ref{tab:reconstruction}; full 1,024-object row included. \\
MeshGPT / PolyGen \citep{meshgpt,polygen} & Clean-CAD mesh-token generation references. & Sequence tokens over mesh geometry, but no released open-world structural state with ownership and repair operations. & Used as mesh-generation context; not treated as evidence for or against the operational structural-state claim. \\
MeshAnythingV2 \citep{meshanything} & Privileged point-normal mesh-generation control on the same 1,024 source objects. & Mesh-only output; no exposed component ownership, attachment validity, or repairable seam state. & With GT point-normal inputs and source-frame mapping: Chamfer 0.0492, Hausdorff 0.4325, contamination 0.1173, separation 0.9292, normal consistency 0.0041. These numbers measure source-frame geometry while Table~\ref{tab:capability_gap} records the missing native operations. \\
MeshArt \citep{meshart} & External mesh-generation reference under native conditioning. & Mesh/remeshing output rather than an inspectable assembly state. & Used as external SOTA context for mesh generation, not as a structural-state comparison. \\
LoST \citep{lost} & Related latent-token perspective. & Does not expose the ownership/seam variables required by our repair tasks in a released same-object interface. & Discussed conceptually as related tokenization work. \\
\bottomrule
\end{tabular}
}
\end{table}

The MeshAnythingV2 comparison is deliberately favorable to a geometry-only generator: we export the fixed 1,024 Objaverse-LVIS objects as dense GT point-normal inputs, reproduce the released inference script's subsampling, centering, and scaling convention, and map each generated mesh back to the source frame before scoring. This removes canonical-frame mismatch as the explanation. The remaining distinction is operational: plausible mesh output still contains no native component ownership, candidate seam variables, or latent operator that can reject and repair an invalid attachment. We therefore treat MeshAnythingV2 as a released-interface capability control rather than an apples-to-apples reconstruction baseline: it answers whether a strong mesh generator's public output exposes the state variables needed by our repair task, not whether it is optimized for our deterministic component-owned evaluator. MeshGPT is not assigned a numeric row for the same reason: the relevant question is whether the released representation exposes the structural operations evaluated here.

\begin{figure}[!tbp]
\centering
\includegraphics[width=0.98\textwidth]{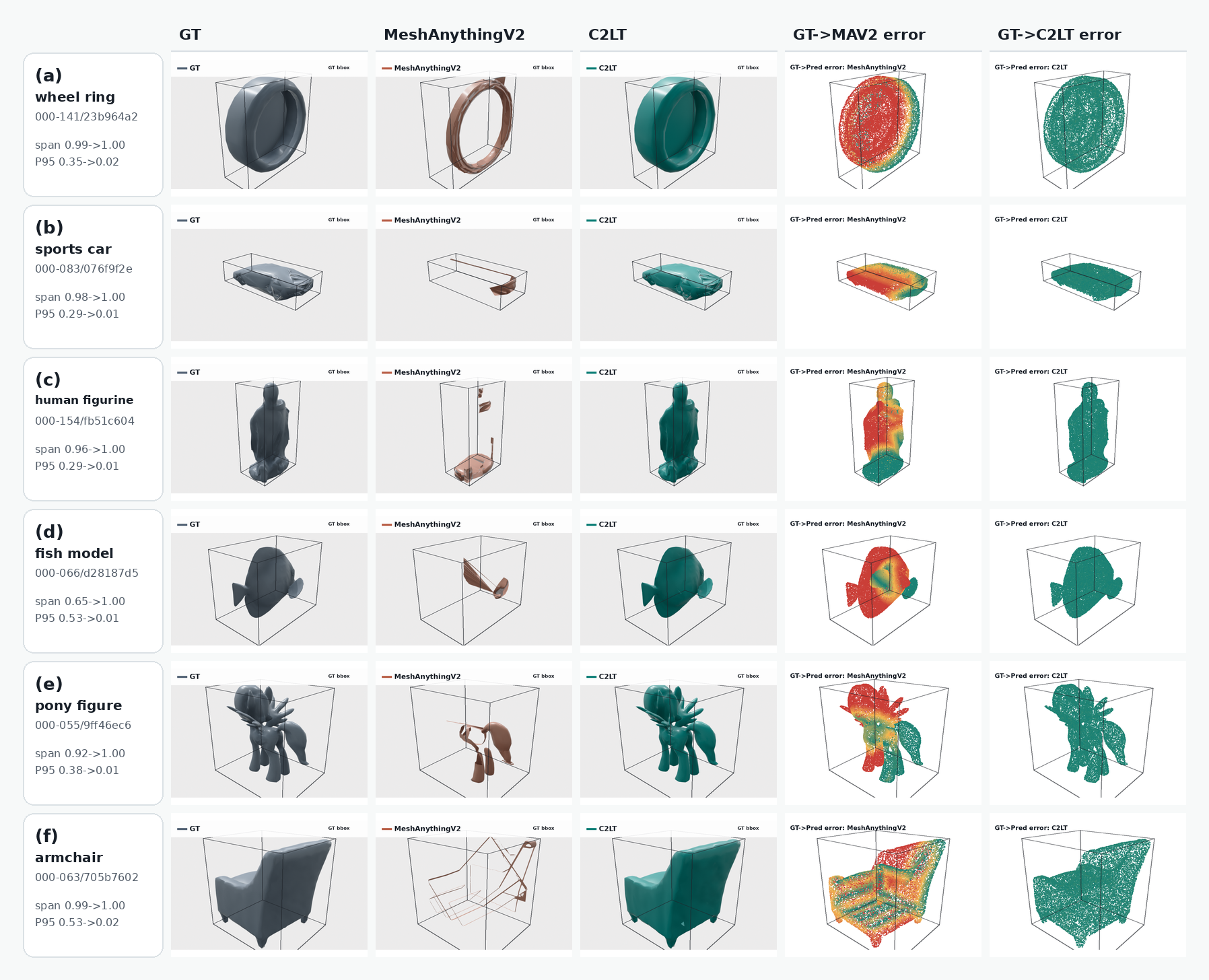}
\caption{\textbf{External MeshAnythingV2 capability-gap evidence.} Each object is shown in the GT reference frame with the GT bounding box overlaid. MeshAnythingV2 receives GT-derived point-normal samples and is mapped back with the released script's center/scale convention; the bottom panels color GT surface points by distance to the source-frame prediction. The comparison shows plausible geometry output under favorable conditioning, but no returned component ownership, attachment variables, or repairable seam state.}
\label{fig:external_meshanythingv2_audit}
\end{figure}

\begin{table}[!tbp]
\centering
\caption{\textbf{Source-frame measurements for the MeshAnythingV2 examples.} Span ratio is measured against the GT bounding-box span in the shared reference frame, GT$\to$Pred P95 is the 95th percentile nearest-surface distance from GT samples to the predicted mesh, and Coverage@0.05 is the fraction of GT samples within distance 0.05 of the prediction. These measurements support the capability-gap reading in Table~\ref{tab:capability_gap}: they evaluate geometry preservation, while the missing ownership and seam-repair variables are representational rather than recoverable from the mesh output alone.}
\label{tab:external_meshanythingv2_diagnostic}
\footnotesize
\resizebox{\textwidth}{!}{\begin{tabular}{llcccccc}
\toprule
Sample & Object & \multicolumn{2}{c}{Span ratio $\uparrow$} & \multicolumn{2}{c}{GT$\to$Pred P95 $\downarrow$} & \multicolumn{2}{c}{Coverage@0.05 $\uparrow$} \\
 & & MeshAnythingV2 & C2LT & MeshAnythingV2 & C2LT & MeshAnythingV2 & C2LT \\
\midrule
(a) & wheel ring & 0.988 & 1.003 & 0.354 & 0.018 & 0.228 & 1.000 \\
(b) & sports car & 0.980 & 1.002 & 0.292 & 0.014 & 0.235 & 1.000 \\
(c) & human figurine & 0.956 & 1.000 & 0.288 & 0.011 & 0.337 & 1.000 \\
(d) & fish model & 0.648 & 1.000 & 0.526 & 0.010 & 0.162 & 1.000 \\
(e) & pony figure & 0.919 & 1.000 & 0.380 & 0.013 & 0.358 & 1.000 \\
(f) & armchair & 0.994 & 0.996 & 0.526 & 0.017 & 0.223 & 1.000 \\
\bottomrule
\end{tabular}
}
\end{table}

\section{Additional State-to-Mesh Gallery}
\label{sec:supp_mesh_gallery}

\begin{figure}[p]
\centering
\includegraphics[width=0.92\textwidth]{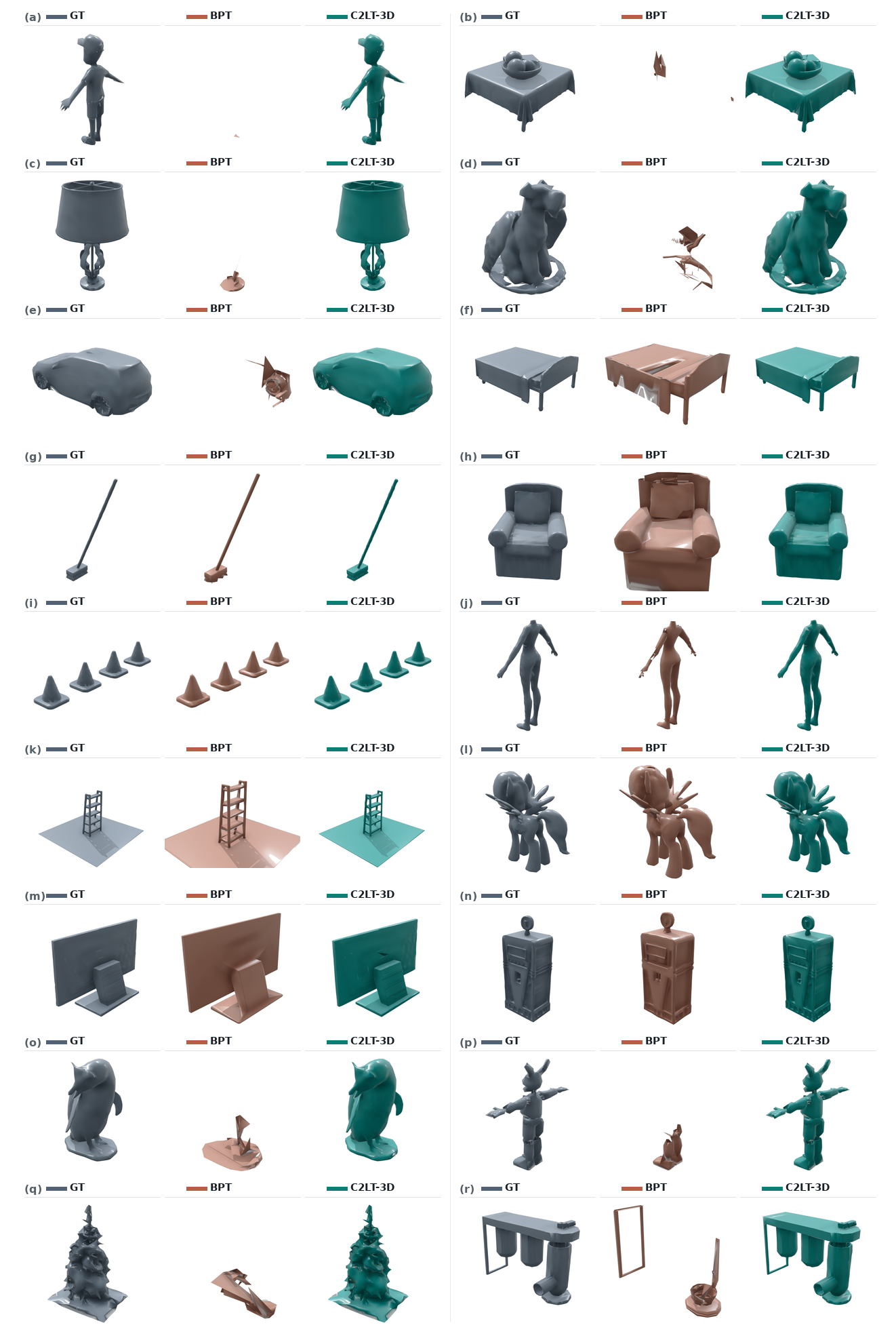}
\caption{\textbf{Explicit triangle-mesh realization from C2LT chart states.} Each block compares the full source mesh, the evaluated BPT mesh, and the C2LT-3D mesh-token realization decoded from chart prefixes. The figure is teacher-forced and therefore should be read as a mesh-realization capacity evaluation rather than an unconditional generation result. Across eighteen additional recognizable open-world structures not used in the main object-level figure, including cases where BPT fragments or drops most of the object, the C2LT realization preserves more complete triangle-mesh structure without post-decoding mesh repair. Table~\ref{tab:explicit_mesh_realization_full} gives the corresponding quantitative summary.}
\label{fig:explicit_mesh_realization}
\end{figure}

Figure~\ref{fig:explicit_mesh_realization} provides the larger non-overlapping gallery for the explicit mesh-token realization study in Section~\ref{sec:supp_explicit_mesh_realization}. The main text keeps the primary GT--BPT--C2LT comparison in Figure~\ref{fig:openworld_object_qualitative} and uses this appendix gallery to show additional recognizable open-world structures without repeating the same qualitative evidence path in the main narrative. Table~\ref{tab:explicit_mesh_realization_full} gives the decoder specification and quantitative summary behind the main-text mesh-realization discussion.

\section{Scope of Local Seam Validity}
\label{sec:supp_hypothesis}

The main text uses seam compatibility as a local structural operator. Here we make its scope explicit: local seam validity is useful for ranking, repair, and constrained decoding, but it is not a formal proof of global topology or loop closure.

\textbf{Observation 1 (Depth-Moderated Risk of Global Collision):} Let $G=(V, E)$ be the structural assembly graph of a generated object. Assume the autoregressive derivation is acyclic with maximum depth $D_{\max}$. If every local edge $e_{ij} \in E$ satisfies a local non-collision threshold (i.e., penetration volume $V_{\mathrm{pen}}(i,j) \le \delta_{\mathrm{coll}}$), and all generated geometries are bounded within a local radius $R_{\max}$, then any non-local collision must arise from accumulated local transform and boundary errors along graph paths, or from violations of the acyclic assumption.

\textbf{Intuitive justification.}
Because $G$ is a tree, the physical relation between any two non-adjacent nodes $t_i$ and $t_k$ is governed by the sequence of local relative transforms $\Delta T$ along the unique path connecting them. If each local attachment satisfies the non-penetration threshold, then a non-local intersection can only appear through accumulated transform error, loose local thresholds, or a path that folds the object back toward an earlier region of space. This observation motivates our use of seam compatibility as a local relational constraint while also making clear why a purely local interface cannot provide a formal global non-intersection guarantee.

\textbf{Where local validity is insufficient.} If the generated object contains structural cycles (for example, a wheel, axle, chassis, and wheel housing that must align around a loop), the acyclic assumption fails. Local seam validity is then insufficient to guarantee global closure without an additional refinement or verification pass.

\section{Benchmark Evaluator Definitions and Metric Formulas}
\label{sec:supp_metrics}

This section gives the exact evaluator used by the main and appendix tables. The first two paragraphs state the evaluation contract, the equations define each metric, and Table~\ref{tab:metric_validity} summarizes why the custom structural scores are always interpreted together with standard geometry and intervention evidence.

\textbf{Metric Sanity and Interpretation.} The structure-sensitive metrics are used as paired measures under a fixed evaluator, not as stand-alone physical measurements. All methods in a table are evaluated on the same object IDs, same query lattice, and same object-adaptive threshold convention; no threshold is tuned per method. We interpret these measures only together with standard geometry measures (Chamfer, Hausdorff, and normal consistency), qualitative object-level views, the high-complexity slice, and the unfiltered Objaverse-LVIS robustness check. Table~\ref{tab:metric_validity} summarizes the robustness and validity checks behind this evaluator. This prevents a method from being favored solely because it optimizes one custom structural score while degrading ordinary geometric fidelity.

\textbf{Table conventions.} For all paired fixed-object comparisons, improvements are computed per object before aggregation. For lower-is-better metrics such as Chamfer, Hausdorff, and contamination, the tabulated improvement is baseline minus C2LT-3D; for higher-is-better metrics such as separation and normal consistency, it is C2LT-3D minus baseline. Object win rate is the fraction of fixed benchmark objects on which the per-object improvement is positive. Bootstrap intervals resample objects or repair tasks, as stated in each caption.

\textbf{Component Separation Score ($S_{\mathrm{sep}}$):} Evaluates the preservation of distinct physical boundaries for intersecting components. Let $\hat{X}_k$ be the predicted point set owned by component $k$, and let $d(x, A)=\min_{a\in A}\|x-a\|_2$. We use an object-adaptive threshold $\tau=\max(0.02\cdot\mathrm{extent},10^{-3})$ and measure how often independently owned predicted components merge into each other:
\begin{equation}
S_{\mathrm{sep}}=1-\frac{1}{|\mathcal{P}|}\sum_{(i,j)\in\mathcal{P}}
\frac{1}{2}\left[
\frac{1}{|\hat{X}_i|}\sum_{x\in \hat{X}_i}\mathbf{1}\{d(x,\hat{X}_j)<\tau\}
+\frac{1}{|\hat{X}_j|}\sum_{x\in \hat{X}_j}\mathbf{1}\{d(x,\hat{X}_i)<\tau\}
\right],
\end{equation}
where $\mathcal{P}$ is the set of nonempty valid predicted component pairs. If no predicted component exists, the evaluator sets $S_{\mathrm{sep}}=0$; if predicted components exist but no pair exists, it assigns the neutral value $S_{\mathrm{sep}}=1$. Higher values indicate less predicted cross-component overlap.

\textbf{Cross-Component Contamination Rate ($C_{\mathrm{rate}}$):} Measures unwanted blending of adjacent parts by mapping each predicted point back to its own ground-truth component support $X_k$ and the nearest support from all other components $X_{\neg k}$. A predicted point owned by $k$ is flagged as contaminated when it is closer to another component than to its own support by more than the same adaptive threshold, or when it lies inside another component's boundary band while being outside its own:
\begin{equation}
C_{\mathrm{rate}}=\frac{\sum_k\sum_{x\in\hat{X}_k}\mathbf{1}\left[
d(x,X_{\neg k})+\tau<d(x,X_k)\;\vee\;\left(d(x,X_{\neg k})<\tau \wedge d(x,X_k)>\tau\right)
\right]}{\sum_k|\hat{X}_k|}.
\end{equation}
Lower values indicate less cross-component leakage. If no predicted component exists, the evaluator sets $C_{\mathrm{rate}}=1$; if predicted components exist but no cross-component comparison is defined, it assigns the neutral value $C_{\mathrm{rate}}=0$. This point-set definition is the one used for Tables~\ref{tab:reconstruction} and~\ref{tab:bpt_subset}.

\textbf{Support Violation Measure:} For auxiliary support analysis, we perform downward ray-casting from every distinct connected component. If a component's rays do not intersect another component or the ground plane within a predefined threshold $\delta_{\mathrm{support}}$, it is flagged as ungrounded. This measure characterizes support violations and is separate from the point-set metrics in Tables~\ref{tab:reconstruction} and~\ref{tab:bpt_subset}.

\paragraph{Additional Metric Definitions for Downstream Tables.}
\textbf{Object-Level Geometry Metrics.} Let $\hat{X}$ be the predicted object-level point set and $X$ the reference support surface. Chamfer is the mean bidirectional nearest-neighbor distance,
\begin{equation}
\mathrm{CD}(\hat{X},X)=\frac{1}{2}\left(\frac{1}{|\hat{X}|}\sum_{\hat{x}\in\hat{X}}d(\hat{x},X)+\frac{1}{|X|}\sum_{x\in X}d(x,\hat{X})\right),
\end{equation}
and Hausdorff is the maximum bidirectional nearest-neighbor distance,
\begin{equation}
\mathrm{HD}(\hat{X},X)=\max\left\{\max_{\hat{x}\in\hat{X}}d(\hat{x},X),\max_{x\in X}d(x,\hat{X})\right\}.
\end{equation}
Empty-set cases are assigned distance 0 when both sets are empty and distance 1 when only one set is empty. Normal Consistency is the mean cosine similarity between predicted and reference query normals over valid normal locations $\Omega$:
\begin{equation}
\mathrm{NC}=\frac{1}{|\Omega|}\sum_{q\in\Omega}
\frac{\hat{n}_q^\top n_q}{\max(\|\hat{n}_q\|_2,10^{-6})\max(\|n_q\|_2,10^{-6})}.
\end{equation}

\textbf{Controlled Serialization Metrics (Appendix Table~\ref{tab:generation}).} Each decoded object is mapped to an 11-dimensional structural feature vector $\phi(o)$ consisting of normalized object extents, normalized covariance eigenvalues, normalized unit count, mean unit extent, centroid spread, component-intersection quality, and boundary clarity. Structural FID follows the Fr\'echet-distance form of FID \citep{heusel2017gans} between Gaussian fits of decoded features $(\mu_g,\Sigma_g)$ and reference features $(\mu_r,\Sigma_r)$:
\begin{equation}
\mathrm{FID}_{\mathrm{struct}}=\|\mu_g-\mu_r\|_2^2+
\mathrm{Tr}\left(\Sigma_g+\Sigma_r-2(\Sigma_g^{1/2}\Sigma_r\Sigma_g^{1/2})^{1/2}\right),
\end{equation}
where the inner square-root term is the principal positive-semidefinite matrix square root in the Gaussian Fr\'echet/Wasserstein-2 distance.
For the local assembly statistics, we construct a $k$-NN graph over generated unit centroids. Let $O_{ij}$ be the symmetric fraction of points from units $i$ and $j$ that fall within $0.12$ times their local scale, and let
\begin{equation}
B_{ij}=\exp\left(-\frac{\mathrm{CD}(B_i,B_j)}{0.15\,s_{ij}}\right)(1-O_{ij})
\end{equation}
where $B_i$ and $B_j$ are boundary samples and $s_{ij}$ is the larger local unit extent. The raw component-intersection and boundary-clarity statistics are
\begin{equation}
IQ_{\mathrm{raw}}=1-\frac{1}{|E_k|}\sum_{(i,j)\in E_k}O_{ij},\qquad
BC_{\mathrm{raw}}=\frac{1}{|E_k|}\sum_{(i,j)\in E_k}B_{ij}.
\end{equation}
The final scores compare these raw statistics with the reference-set means:
\begin{equation}
IQ=\exp(-|IQ_{\mathrm{raw}}-\mu_{IQ}^{\mathrm{ref}}|/0.15),\qquad
BC=\exp(-|BC_{\mathrm{raw}}-\mu_{BC}^{\mathrm{ref}}|/0.05).
\end{equation}
Higher values therefore indicate more reference-like non-colliding part interaction and boundary behavior, rather than merely maximizing separation at all costs.

\textbf{Seam Discrimination Metrics (Table~\ref{tab:seam_metrics}).} AUC and AP evaluate how well seam scores rank valid candidate seams above invalid ones under the evaluator's binary seam-validity threshold. Collision Brier is
\begin{equation}
\mathrm{Brier}_{\mathrm{coll}}=\frac{1}{N}\sum_{e=1}^N(\hat{p}^{\mathrm{coll}}_e-y^{\mathrm{coll}}_e)^2.
\end{equation}
Top-1 Precision and Top-3 Recall are computed per source chart over its candidate outgoing seams, then averaged across sources, so they measure whether the highest-ranked local seam proposals recover valid attachments.

\textbf{Repair Metrics (Table~\ref{tab:repair_metrics} and Table~\ref{tab:repair_main} in the main text).} Let $r_m$ be the reference parent for repair task $m$, $V_m$ the set of valid candidate parents, and $\pi_m(k)$ the $k$-th ranked candidate parent. Valid Repair@1 and Exact Parent@1 are
\begin{equation}
\mathrm{Valid@1}=\frac{1}{M}\sum_{m=1}^M\mathbf{1}\{\pi_m(1)\in V_m\},\quad
\mathrm{Parent@1}=\frac{1}{M}\sum_{m=1}^M\mathbf{1}\{\pi_m(1)=r_m\}.
\end{equation}
For the extended ranking table, Valid@$K$ and Parent@$K$ replace $\pi_m(1)$ with the top-$K$ set. Valid MRR and Exact MRR use the reciprocal rank of the first valid parent and of $r_m$, respectively:
\begin{equation}
\mathrm{MRR}_{\mathrm{valid}}=\frac{1}{M}\sum_{m=1}^M \frac{1}{\rho_m^{\mathrm{valid}}},\quad
\mathrm{MRR}_{\mathrm{parent}}=\frac{1}{M}\sum_{m=1}^M \frac{1}{\rho_m^{\mathrm{parent}}},
\end{equation}
where $\rho_m^{\mathrm{valid}}=\min\{k:\pi_m(k)\in V_m\}$ and $\rho_m^{\mathrm{parent}}=\min\{k:\pi_m(k)=r_m\}$; if no such candidate appears in the ranked list, we set $\rho=\infty$ so the reciprocal-rank contribution is zero.
Confidence intervals in Table~\ref{tab:repair_rank_ci} are 95\% bootstrap intervals over repair tasks using 5,000 resamples. The hard subset contains edge-target tasks on which the nearest-neighbor heuristic prefers an invalid parent to the valid reference.

\textbf{Inference-only repair policy.} For the serialized-prefix analysis in Table~\ref{tab:intervention_stress}, the repair policy ranks candidate parents using only inference-time quantities available to the model: the learned seam score, a local distance prior, and predicted collision/invalid risk. For candidate $k$ within a repair task, let tildes denote min--max normalization over that task's candidate set. The policy is
\begin{equation}
S_{\mathrm{repair}}(k)=\tilde{s}_{\psi}(k)+0.5\,\tilde{s}_{\mathrm{dist}}(k)-0.5\,\tilde{p}_{\mathrm{coll}}(k)-0.5\,\tilde{p}_{\mathrm{invalid}}(k).
\end{equation}
It deliberately excludes support-overlap, boundary-Chamfer, and normal-consistency target features, which are useful geometric verifier measures but too close to the supervision signal for the main repair comparison.

\textbf{Canonicalization Ablation Metrics (Table~\ref{tab:canonicalization_ablation}).} \emph{Geo PPL} and \emph{Bnd PPL} are the empirical perplexities of the geometry-token and boundary-token distributions over the validation charts; \emph{Geo Util.} and \emph{Bnd Util.} are the corresponding active-code fractions relative to the full codebook size. \emph{Occ. BCE} is the mean binary occupancy cross-entropy on the local query lattice. \emph{Patch IoU} is the mean intersection-over-union between thresholded occupied queries and ground-truth occupied queries within each chart. \emph{Normal Cosine} is the mean cosine similarity between predicted and ground-truth query normals on valid normal locations. These are local-field metrics and should not be confused with the object-level open-world reconstruction metrics above.

\textbf{Systems and Mesh-Realization Metrics.} Decode time per object is total measured wall-clock divided by the number of evaluated objects. For preprocessing tables, throughput is $N/T$ objects per second, projected split cost is $N_{split}/(N/T)$, and mean end-to-end time per object is $T/N$ on the sampled subset. In the explicit mesh-token realization table, cross-entropy is the teacher-forced next-token loss,
\begin{equation}
\mathrm{CE}=-\frac{1}{L}\sum_{\ell=1}^{L}\log p_\theta(y_\ell\mid y_{<\ell},\mathrm{C2LT\ prefix}),
\end{equation}
perplexity is $\exp(\mathrm{CE})$, and token accuracy is the fraction of teacher-forced positions whose argmax token matches the target. Mesh-realization Chamfer is computed between decoded mesh samples and the reference chart-token union or source surface stated in the table; p95 is the 95th percentile of the corresponding per-object Chamfer values.

\begin{table}[!tbp]
\centering
\caption{\textbf{Metric robustness and validity evidence.} The structural metrics are not treated as isolated scalar scores. Each is paired with standard geometry measures, fixed-object paired statistics, or a task-level intervention check that tests whether the measured failure mode is actionable.}
\label{tab:metric_validity}
\small
\setlength{\tabcolsep}{4pt}
\renewcommand{\arraystretch}{1.18}
\begin{tabular}{@{}P{0.20\textwidth}|P{0.35\textwidth}P{0.37\textwidth}@{}}
\toprule
\textbf{Metric family} & \textbf{What it validates} & \textbf{Robustness / linked evidence} \\
\midrule
Chamfer and Hausdorff & Standard object-level geometric fidelity and worst-case surface deviation. & Same 1,024 object IDs and query lattice; paired object bootstrap and win rates remain positive against both BPT and VQ-Patch in Tables~\ref{tab:bpt_subset} and~\ref{tab:matched_bootstrap}. \\
Normal consistency & Local surface orientation quality, guarding against methods that improve distance while degrading surface detail. & Presented beside distance and structural metrics rather than optimized alone; paired bootstrap intervals are strictly positive in Table~\ref{tab:matched_bootstrap}. \\
Contamination and cross-component leakage & Cross-component blending, ownership leakage, and false support transfer between neighboring parts. & Fixed object-adaptive threshold convention shared by all methods; no threshold is tuned per method. Gains persist on the fixed 1,024-object set and the high-complexity slice in Tables~\ref{tab:bpt_subset}, \ref{tab:matched_bootstrap}, and~\ref{tab:complexity_slice}. \\
Separation & Whether distinct components remain structurally isolated instead of being fused into a single support. & Interpreted jointly with Chamfer, Hausdorff, contamination, and normal consistency, so a method cannot win by simply pushing components apart or eroding geometry. \\
Repair and seam-ranking metrics & Whether the latent state exposes attachment variables that can be acted on when local geometry selects an invalid parent. & 1,964-task inter-part edge bank with hard and heuristic-fail subsets, Top-$k$/MRR, and bootstrap CIs in Tables~\ref{tab:repair_rank_ci}, \ref{tab:repair_metrics}, and~\ref{tab:intervention_stress}; dense support is only an external verifier control. \\
\bottomrule
\end{tabular}
\end{table}

\begin{figure}[!tbp]
\centering
\includegraphics[width=0.98\textwidth]{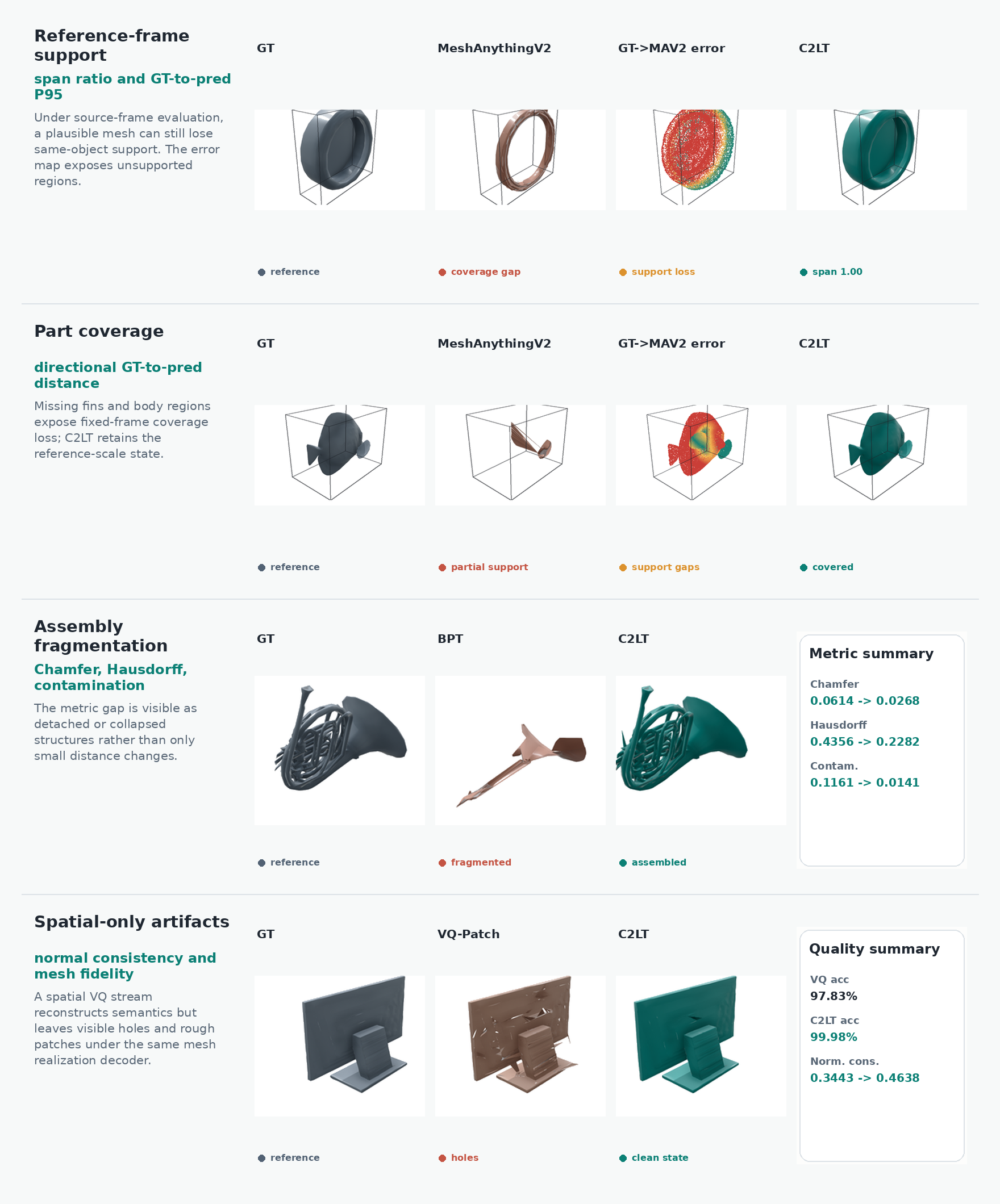}
\caption{\textbf{Visual validation for structure-sensitive metrics.} Each row pairs a metric family with the visible failure mode it is intended to capture: source-frame support loss, missing part coverage under fixed-frame evaluation, assembly fragmentation, and spatial-only mesh artifacts. The figure is not an additional benchmark; it connects the metric interpretation to representative fixed-object evidence beside the corresponding aggregate quantities.}
\label{fig:structural_metric_witness}
\end{figure}

\section{Unfiltered Objaverse-LVIS Robustness Check}
\label{sec:supp_raw_objaverse}

The main text uses the geometry-clean Objaverse-LVIS split as the controlled benchmark. We also evaluate a harder unfiltered Objaverse-LVIS split obtained from the same normalized source root but without the geometry-clean filtering step. This keeps the data source fixed while isolating the effect of filtering itself. We evaluate 5,000 unseen objects sampled from the same source distribution; Table~\ref{tab:raw_openworld} gives the resulting coarse metrics.

\begin{table}[!tbp]
\centering
\caption{\textbf{Unfiltered Objaverse-LVIS robustness check.} The benchmark is derived from the same source root as the geometry-clean Objaverse-LVIS split, but without geometry-clean filtering.}
\label{tab:raw_openworld}
\resizebox{0.8\textwidth}{!}{
\begin{tabular}{l|cccc}
\toprule
\textbf{Method} & \textbf{Chamfer $\downarrow$} & \textbf{Contamination $\downarrow$} & \textbf{Separation $\uparrow$} & \textbf{Norm. Cons. $\uparrow$} \\
\midrule
C2LT-3D Seam Head Only & 0.0370 & 0.0622 & 0.9671 & 0.3797 \\
C2LT-3D Learned Seam Prior & 0.0384 & 0.0619 & 0.9675 & 0.3799 \\
C2LT-3D Component-Owned Realization & 0.0370 & 0.0622 & 0.9671 & 0.3797 \\
\bottomrule
\end{tabular}
}
\end{table}

These unfiltered-set results closely track the geometry-clean benchmark, suggesting that the structural evaluator measures a similar coarse reconstruction regime in both settings. For this reason, we keep the geometry-clean Objaverse-LVIS split as the main benchmark and use the unfiltered split as complementary robustness evidence rather than as a second main table.

\section{Fixed Objaverse-LVIS Robustness Evaluation}
\label{sec:supp_bpt_subset}

This section provides the full 1,024-object comparison underlying Table~\ref{tab:reconstruction}, including Hausdorff distance and decode time. The objects are sampled once from the geometry-clean filtered split and then fixed before evaluation, so BPT, VQ-Patch, and C2LT-3D are measured on identical asset IDs under the same query-lattice structural protocol.

This fixed-object benchmark is the performance comparison used for reconstruction claims. The separate released-interface analysis in Table~\ref{tab:external_sota_audit} asks whether mesh-generation interfaces expose ownership and seam-repair state as native operations.

\begin{table}[!tbp]
\centering
\caption{\textbf{1,024-object Objaverse-LVIS comparison.} Boldface highlights the best quality scores and fastest decode time.}
\label{tab:bpt_subset}
\resizebox{0.98\textwidth}{!}{
\begin{tabular}{lccccccc}
\toprule
\textbf{Method} & \textbf{Subset Size} & \textbf{Chamfer $\downarrow$} & \textbf{Hausdorff $\downarrow$} & \textbf{Contamination $\downarrow$} & \textbf{Separation $\uparrow$} & \textbf{Norm. Cons. $\uparrow$} & \textbf{Decode Time} \\
\midrule
BPT (500M pretrained model) & 1024 & 0.0614 & 0.4356 & 0.1161 & 0.9220 & 0.0569 & 89,519s total ($\sim$87.42s/object) \\
VQ-Patch (Spatial-Only) & 1024 & 0.0303 & 0.3277 & 0.0727 & 0.9608 & 0.3443 & \textbf{0.90s total} ($\sim$0.0009s/object) \\
\textbf{C2LT-3D (Ours)} & 1024 & \textbf{0.0268} & \textbf{0.2282} & \textbf{0.0141} & \textbf{0.9780} & \textbf{0.4638} & 34.05s total ($\sim$0.0332s/object) \\
\bottomrule
\end{tabular}
}
\end{table}

These fixed-set values provide the detailed counterpart to the main reconstruction table. VQ-Patch is the fastest spatial-compression control, while C2LT-3D has better values on all five quality metrics; BPT remains substantially more expensive at 89,519 seconds of total decoding time. This table evaluates component-owned object realization, while the seam prior is evaluated most directly in the repair and intervention benchmarks.

\FloatBarrier

\begin{center}
\centering
\captionof{table}{\textbf{Paired fixed-object bootstrap for the 1,024-object comparison.} Improvements are computed per object on the identical asset IDs used in Table~\ref{tab:bpt_subset}. Positive values always favor C2LT-3D: for distance/error metrics this is baseline minus C2LT-3D, and for higher-is-better metrics this is C2LT-3D minus baseline. Confidence intervals are 95\% bootstrap intervals over objects.}
\label{tab:matched_bootstrap}
\scriptsize
\setlength{\tabcolsep}{3pt}
\resizebox{0.94\textwidth}{!}{
\begin{tabular}{l|l|ccc}
\toprule
\textbf{Baseline} & \textbf{Metric} & \textbf{Mean Improvement} & \textbf{95\% CI} & \textbf{Object Win Rate} \\
\midrule
VQ-Patch & Chamfer & 0.0035 & [0.0033, 0.0038] & 84.8\% \\
VQ-Patch & Hausdorff & 0.0995 & [0.0932, 0.1055] & 88.0\% \\
VQ-Patch & Contamination & 0.0586 & [0.0569, 0.0604] & 99.5\% \\
VQ-Patch & Separation & 0.0173 & [0.0161, 0.0184] & 90.9\% \\
VQ-Patch & Normal Consistency & 0.1195 & [0.1143, 0.1247] & 96.9\% \\
\midrule
BPT & Chamfer & 0.0346 & [0.0301, 0.0393] & 81.2\% \\
BPT & Hausdorff & 0.2073 & [0.1883, 0.2260] & 80.6\% \\
BPT & Contamination & 0.1020 & [0.0993, 0.1047] & 99.2\% \\
BPT & Separation & 0.0561 & [0.0534, 0.0589] & 99.3\% \\
BPT & Normal Consistency & 0.4069 & [0.3895, 0.4243] & 99.7\% \\
\bottomrule
\end{tabular}
}
\end{center}

This paired analysis is a robustness control: object-level resampling tests whether the gains are stable across the fixed benchmark rather than concentrated in a few favorable examples. The intervals remain strictly positive for all five quality metrics against both baselines.

\begin{table}[!tbp]
\centering
\caption{\textbf{Factorized realization fairness control.} We separate the learned state from the deterministic component-owned realization map on the same fixed 1,024-object protocol. VQ-Patch is granted the same realization policy as an oracle control, where ownership is supplied by the evaluator rather than exposed by the spatial tokens. C2LT-3D is also evaluated with the component-owned filter disabled. The comparison shows that the C2LT state improves distance and normal metrics even without the filter, while component-owned realization specifically suppresses cross-component leakage and remains native to C2LT rather than externally supplied.}
\label{tab:same_filter_control}
\scriptsize
\setlength{\tabcolsep}{3pt}
\resizebox{0.98\textwidth}{!}{
\begin{tabular}{l|c|ccccc}
\toprule
\textbf{Method} & \textbf{Native ownership/seam state?} & \textbf{Chamfer $\downarrow$} & \textbf{Hausdorff $\downarrow$} & \textbf{Contam. $\downarrow$} & \textbf{Separation $\uparrow$} & \textbf{Norm. Cons. $\uparrow$} \\
\midrule
VQ-Patch (Spatial-Only) & No & 0.0303 & 0.3277 & 0.0727 & 0.9608 & 0.3443 \\
VQ-Patch + Same Realization Control & No; evaluator-supplied ownership & 0.0299 & 0.3074 & 0.0116 & 0.9838 & 0.3443 \\
C2LT-3D w/o Component-Owned Realization & Yes; state exposed, filter disabled & 0.0273 & 0.2421 & 0.0788 & 0.9528 & 0.4638 \\
C2LT-3D (Ours) & Yes; exposed ownership and seam state & 0.0268 & 0.2282 & 0.0141 & 0.9780 & 0.4638 \\
\bottomrule
\end{tabular}
}
\end{table}

This control is intentionally conservative for C2LT-3D. Giving VQ-Patch oracle ownership strongly improves contamination and separation, confirming that component ownership is the right variable for part isolation. Disabling component-owned realization in C2LT-3D has the complementary effect: distance and normal metrics remain stronger than VQ-Patch, but contamination rises because cross-component ownership is no longer enforced during realization. The full C2LT-3D row combines both effects and, unlike the oracle VQ-Patch control, also provides attachment validation and latent seam repair. We therefore interpret Table~\ref{tab:reconstruction} together with Table~\ref{tab:repair_main}: the core claim is not that a post-hoc filter is unavailable, but that C2LT-3D makes ownership and seam variables native, inspectable, and actionable.

\begin{table}[!tbp]
\centering
\caption{\textbf{Deterministic realization sensitivity.} We sweep the two fixed rule parameters in the component-owned realization map on a 64-object slice from the fixed 1,024-object evaluation pool. The main setting uses ownership margin $m=0.00$ and a 90\% per-component keep floor. The sweep is not used for model selection; it verifies that the quality gain comes from enforcing component ownership, not from a single fragile threshold.}
\label{tab:realization_sensitivity}
\scriptsize
\setlength{\tabcolsep}{3pt}
\resizebox{0.98\textwidth}{!}{
\begin{tabular}{l|cc|ccccc}
\toprule
\textbf{Policy} & \textbf{Margin} & \textbf{Keep Floor} & \textbf{Chamfer $\downarrow$} & \textbf{Hausdorff $\downarrow$} & \textbf{Contam. $\downarrow$} & \textbf{Separation $\uparrow$} & \textbf{Norm. Cons. $\uparrow$} \\
\midrule
No component-owned filter & -- & -- & 0.0254 & 0.2256 & 0.0766 & 0.9377 & 0.4780 \\
Keep-floor sweep & 0.00 & 0.85 & \textbf{0.0247} & \textbf{0.2060} & \textbf{0.0061} & \textbf{0.9764} & 0.4780 \\
Main setting & 0.00 & 0.90 & 0.0250 & 0.2149 & 0.0141 & 0.9692 & 0.4780 \\
Keep-floor sweep & 0.00 & 0.95 & 0.0253 & 0.2217 & 0.0360 & 0.9576 & 0.4780 \\
Margin sweep & 0.25 & 0.90 & 0.0250 & 0.2155 & 0.0144 & 0.9668 & 0.4780 \\
Margin sweep & 0.50 & 0.90 & 0.0250 & 0.2156 & 0.0164 & 0.9637 & 0.4780 \\
\bottomrule
\end{tabular}
}
\end{table}

The no-ownership row has similar geometry distance but much worse cross-component contamination, which is the failure mode the component-owned realization is designed to suppress. Across the ownership-margin and keep-floor sweeps, the structural metrics remain in the same regime as the main setting. We therefore keep the middle policy fixed for all main evaluations instead of tuning realization parameters per method or per object.

\begin{table}[!tbp]
\centering
\caption{\textbf{High-complexity Objaverse-LVIS slice.} We isolate 54 fixed objects whose inferred partition count is at least 10 and evaluate them with the same structural evaluator used in Table~\ref{tab:reconstruction}. The performance gap widens on this structurally heavier slice: BPT incurs substantially higher contamination and lower separation, while the learned seam-prior C2LT-3D variant retains the strongest structural isolation.}
\label{tab:complexity_slice}
\resizebox{0.82\textwidth}{!}{
\begin{tabular}{lccccc}
\toprule
\textbf{Method} & \textbf{Subset Size} & \textbf{Chamfer $\downarrow$} & \textbf{Contamination $\downarrow$} & \textbf{Separation $\uparrow$} & \textbf{Norm. Cons. $\uparrow$} \\
\midrule
BPT (Spatial Comp.) & 54 & 0.0607 & 0.1341 & 0.9026 & 0.0194 \\
C2LT-3D w/o Partition-Conditioned Context & 54 & 0.0870 & 0.1215 & 0.9431 & 0.2667 \\
C2LT-3D Seam Head Only & 54 & \textbf{0.0359} & 0.0641 & 0.9596 & 0.2446 \\
C2LT-3D Learned Seam Prior & 54 & 0.0386 & \textbf{0.0608} & \textbf{0.9727} & \textbf{0.2477} \\
\bottomrule
\end{tabular}
}
\end{table}

\begin{table}[!tbp]
\centering
\caption{\textbf{Full 5,000-object geometry-clean Objaverse-LVIS benchmark.} We include coarse structure-sensitive metrics on the full filtered benchmark for completeness. Hausdorff distance and decode-time columns are omitted because this full-scale benchmark was not measured under the same timing protocol used in Table~\ref{tab:reconstruction}.}
\label{tab:filtered_full5000}
\resizebox{0.82\textwidth}{!}{
\begin{tabular}{lcccc}
\toprule
\textbf{Checkpoint} & \textbf{Chamfer $\downarrow$} & \textbf{Contamination $\downarrow$} & \textbf{Separation $\uparrow$} & \textbf{Norm. Cons. $\uparrow$} \\
\midrule
C2LT-3D Seam Head Only & 0.0369 & 0.0620 & 0.9670 & 0.3850 \\
C2LT-3D Learned Seam Prior & 0.0383 & 0.0618 & 0.9674 & 0.3848 \\
C2LT-3D Component-Owned Realization & 0.0369 & 0.0620 & 0.9670 & 0.3850 \\
\bottomrule
\end{tabular}
}
\end{table}

\section{Complexity, Compute, and Latency}
\label{sec:supp_complexity}

We characterize the computational cost of the interface-centric generative state, separating offline corpus construction from online object decoding.

\begin{table}[!tbp]
\centering
\caption{\textbf{Measured Compute and Latency Budget.} All numbers are measured wall-clock on a single NVIDIA GeForce RTX 5090. Training times include validation passes. When a training phase uses multiple continuation schedules, the table gives the runtime of the model instance used in the main-text evaluations. Representative subset preprocessing measurements are provided in Table~\ref{tab:preprocess_budget}; here we list corpus scale rather than duplicating a second timing table inside the main budget summary.}
\label{tab:compute}
\resizebox{0.98\textwidth}{!}{
\begin{tabular}{lcc}
\toprule
\textbf{Component} & \textbf{Measured Budget} & \textbf{Notes} \\
\midrule
Preprocessed ShapeNet corpus & 49,518 objects / 3.17M charts & 44,473-object train pool plus 2,535 validation and 2,510 held-out test objects; representative subset timings in Table~\ref{tab:preprocess_budget} \\
Preprocessed geometry-clean Objaverse-LVIS corpus & 42,100 objects / 2.69M charts & geometry-clean subset used in open-world evaluation \\
Tokenizer training & 20.36 h & 52 realized train epochs, batch size 256 \\
Contextualizer final continuation & 1.92 h & 60 train epochs, batch size 128 \\
Seam-prior head-only training & 0.86 h & 32 realized train epochs, batch size 128 \\
Repair-focused seam-prior continuation & 0.57 h & 20 train epochs, batch size 128 \\
\midrule
VQ-Patch decode & 0.88 ms / object & 1,024-object Objaverse-LVIS set, object decode only \\
C2LT-3D component-owned realization & 33.25 ms / object & object decoding on the 1,024-object set \\
BPT (500M) generation & 87.42 s / object & 1,024-object Objaverse-LVIS set, mesh generation only \\
BPT (500M) full evaluation & 89.72 s / object & generation + structure metrics on the 1,024-object set \\
\bottomrule
\end{tabular}
\vspace{0.05em}
}
\end{table}
Table~\ref{tab:compute} gives measured wall-clock rather than provisional estimates. VQ-Patch is fastest because it uses the lightest spatial-only realization path, while the component-owned C2LT-3D object decoder runs in 33.25 ms/object and remains roughly three orders of magnitude faster than BPT generation. The seam-prior contribution is evaluated primarily through attachment ranking and repair, where its relational variables are directly exercised.

\begin{table}[!tbp]
\centering
\caption{\textbf{Representative preprocessing budget on sampled subsets.} We measure the actual mesh-to-chart-to-graph pipeline on 128 sampled objects per dataset with 8 workers, then give the observed throughput and a coarse corpus-scale projection over the measured split. These measurements quantify the preprocessing burden without assuming a fully instrumented monolithic build of the entire storage pipeline.}
\label{tab:preprocess_budget}
\resizebox{0.98\textwidth}{!}{
\begin{tabular}{l|cccccc}
\toprule
\textbf{Dataset Split} & \textbf{Sampled Objects} & \textbf{Workers} & \textbf{Wall-Clock} & \textbf{Throughput} & \textbf{Projected Split Cost} & \textbf{Mean End-to-End / Object} \\
\midrule
ShapeNet train pool & 128 & 8 & 65.0s & 1.97 obj/s & 6.27 h over 44,473 objects & 3.96 s \\
Geometry-clean Objaverse-LVIS test & 128 & 8 & 80.9s & 1.58 obj/s & 7.39 h over 42,100 objects & 4.92 s \\
\bottomrule
\end{tabular}
}
\end{table}

The per-object breakdown shows that preprocessing cost is dominated by canonical chart construction and graph/seam construction rather than by mesh loading or file I/O. On ShapeNet, the mean object requires 2.07s for canonical chart construction and 1.80s for graph/seam construction; on the geometry-clean Objaverse-LVIS split, the corresponding means are 3.13s and 1.66s. These measurements do not eliminate the preprocessing-cost limitation, but they make the systems burden concrete: the dominant expense lies in local canonical chart building and subsequent seam graph construction, not in the millisecond-scale decode path used at evaluation time. In other words, the main systems cost is now measured at the representative-subset level even though a single fully instrumented corpus-wide build measurement remains future work.

\begin{table}[!tbp]
\centering
\caption{\textbf{Representative lifecycle cost for the main evaluation pipelines.} We separate amortized offline corpus construction from online inference-time use. For C2LT-3D, the offline build numbers are the representative subset preprocessing measurements from Table~\ref{tab:preprocess_budget}, and the training path sums the tokenizer, context, and seam-head-only runtimes used by the main reconstruction pipeline. For the released BPT baseline, our evaluation setup exposes measured online generation and full-evaluation latency, but not a separately instrumented offline preprocessing stage or a directly comparable retraining run.}
\label{tab:lifecycle_cost}
\small
\setlength{\tabcolsep}{4pt}
\renewcommand{\arraystretch}{1.18}
\begin{tabular}{@{}P{0.22\textwidth}|P{0.37\textwidth}P{0.34\textwidth}@{}}
\toprule
\textbf{Pipeline} & \textbf{Offline construction / training} & \textbf{Online use / evaluation} \\
\midrule
C2LT-3D component-owned realization & Offline build: 6.27 h over the ShapeNet train pool and 7.39 h over geometry-clean Objaverse-LVIS test. Training path: 23.14 h total. & Decode: 33.25 ms / object. Metric evaluation uses the shared structural evaluator summarized in Table~\ref{tab:compute}. \\
BPT 500M baseline & Offline build not exposed by the released pipeline. We reuse the external pretrained baseline for evaluation. & Generation: 87.42 s / object. Full evaluation: 89.72 s / object. \\
\bottomrule
\end{tabular}
\end{table}

Table~\ref{tab:lifecycle_cost} clarifies the systems interpretation: C2LT-3D pays an offline corpus-construction cost that is amortized over downstream decoding, repair, and validation queries, while online object decoding is millisecond-scale.

\section{Mechanistic Evidence for the Relational Seam Prior}
\label{sec:supp_seam_prior}
\begin{figure}[!tbp]
\centering
\includegraphics[width=0.98\textwidth]{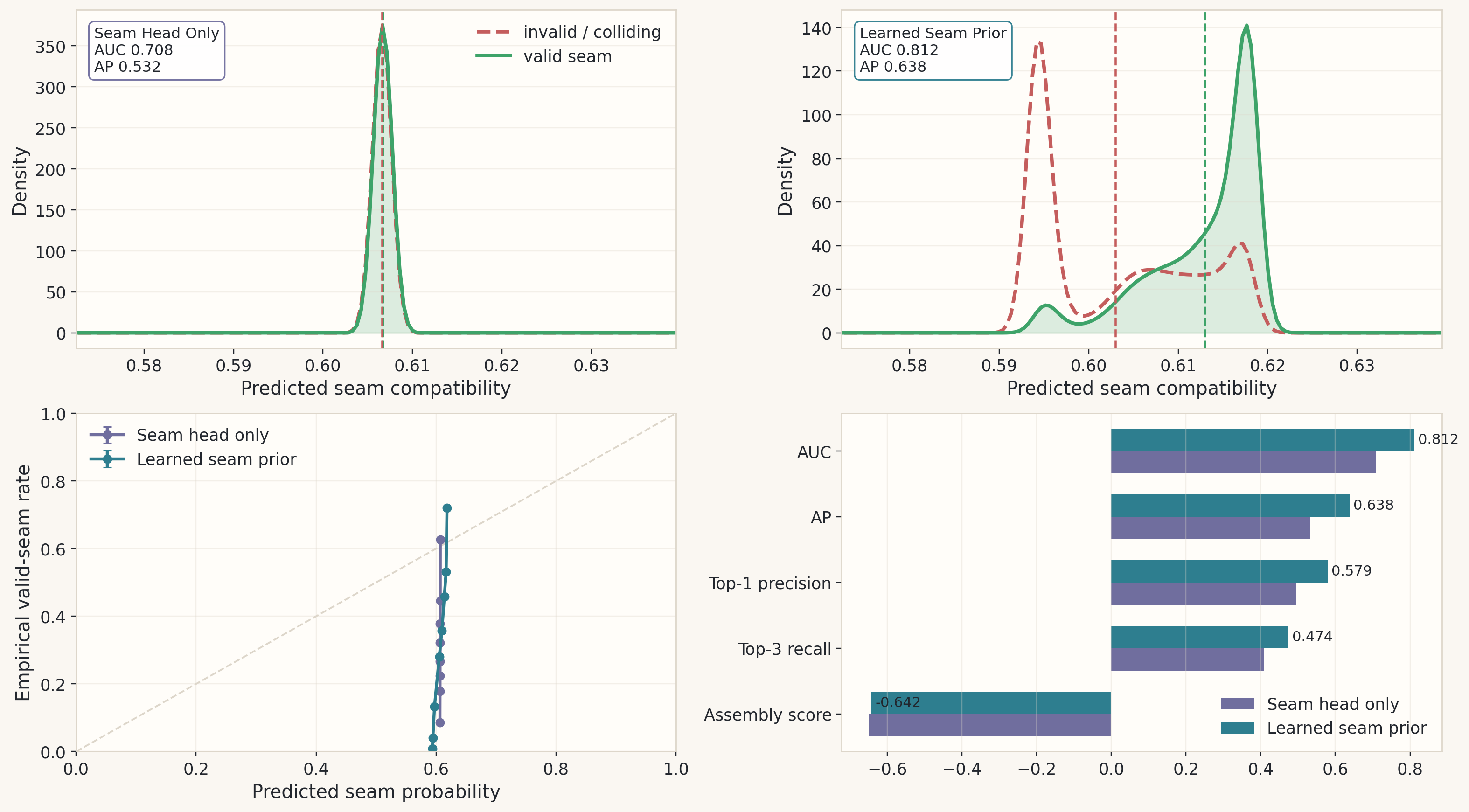}
\caption{\textbf{Mechanistic evidence for the relational seam prior.} Top: score distributions for valid seams versus invalid or colliding candidates under the seam-head-only and learned seam-prior variants. Bottom left: calibration of predicted seam probability against the empirical valid-seam rate. Bottom right: the learned seam prior improves the ranking-oriented seam measures that govern repair, including AUC, average precision, top-1 precision, and top-3 recall.}
\label{fig:seam_histogram}
\end{figure}

We provide mechanistic evidence that the relational seam prior functions as an active assembly rule rather than a passive compatibility regressor.
Figure~\ref{fig:seam_histogram} visualizes this mechanism through score separation, calibration, and ranking-oriented seam measures; Table~\ref{tab:seam_metrics} gives the corresponding numeric discrimination results.

\begin{table}[!tbp]
\centering
\caption{\textbf{Seam discrimination on the ShapeNet validation split.} The learned seam prior improves seam discrimination and ranking quality relative to the seam-head-only variant, even though the main open-world reconstruction table evaluates the shared component-owned object realization.}
\label{tab:seam_metrics}
\resizebox{0.88\textwidth}{!}{
\begin{tabular}{l|ccccc}
\toprule
\textbf{Checkpoint} & \textbf{AUC $\uparrow$} & \textbf{AP $\uparrow$} & \textbf{Collision Brier $\downarrow$} & \textbf{Top-1 Prec. $\uparrow$} & \textbf{Top-3 Recall $\uparrow$} \\
\midrule
C2LT-3D Seam Head Only & 0.7084 & 0.5320 & 0.00131 & 0.4962 & 0.4086 \\
C2LT-3D Learned Seam Prior & \textbf{0.8122} & \textbf{0.6380} & \textbf{0.00126} & \textbf{0.5794} & \textbf{0.4743} \\
\bottomrule
\end{tabular}
}
\end{table}

The learned seam prior strengthens the quantities that govern neighbor retrieval, collision-sensitive ranking, and serialized assembly selection. Its small degradation on coarse open-world reconstruction metrics is why we treat seam adaptation as a complementary relational analysis rather than a replacement for deterministic object realization.

\begin{table}[!tbp]
\centering
\caption{\textbf{Serialized-prefix repair benchmark on the Objaverse-LVIS evaluation set.} In addition to the main edge-bank benchmark, we derive 153 inter-part repair tasks from serialized assembly prefixes. The \emph{hard subset} contains the 51 tasks for which the nearest-neighbor heuristic assigns a higher score to an invalid attachment than to the valid reference.}
\label{tab:repair_metrics}
\resizebox{0.95\textwidth}{!}{
\begin{tabular}{l|cc|cc}
\toprule
\textbf{Method} & \multicolumn{2}{c|}{\textbf{All Inter-Part Tasks (153)}} & \multicolumn{2}{c}{\textbf{Hard Subset (51)}} \\
\cmidrule(lr){2-3}\cmidrule(lr){4-5}
 & \textbf{Valid Repair@1 $\uparrow$} & \textbf{Exact Parent@1 $\uparrow$} & \textbf{Valid Repair@1 $\uparrow$} & \textbf{Exact Parent@1 $\uparrow$} \\
\midrule
Nearest-Neighbor Heuristic & 0.7124 & 0.5686 & 0.1373 & 0.0000 \\
C2LT-3D Seam Head Only & 0.4248 & 0.3725 & 0.3922 & 0.3529 \\
C2LT-3D w/o Partition & 0.4379 & 0.3529 & 0.4118 & 0.2745 \\
C2LT-3D Learned Seam Prior & 0.6275 & 0.5556 & 0.5294 & 0.4902 \\
\textbf{C2LT-3D Repair Policy} & \textbf{0.7451} & \textbf{0.6667} & \textbf{0.6667} & \textbf{0.6275} \\
\bottomrule
\end{tabular}
}
\end{table}

\begin{figure}[!tbp]
\centering
\includegraphics[width=\textwidth]{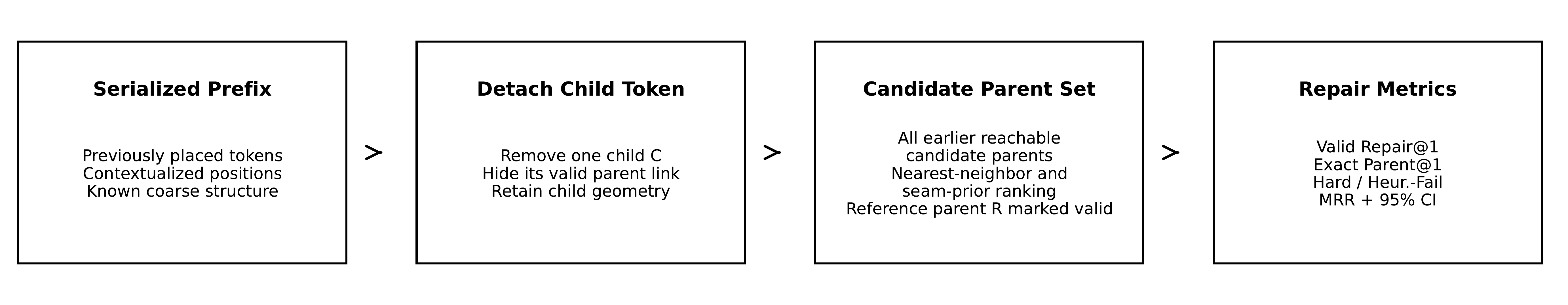}
\caption{\textbf{Construction of the serialized-prefix repair benchmark.} Each prefix task is formed by detaching one serialized child token from a partial object prefix, preserving the earlier tokens as candidate parents, and asking the model to recover the valid parent and attachment relation in latent space. The main text uses the larger inter-part edge-bank benchmark; this figure documents the complementary prefix protocol used in Table~\ref{tab:repair_metrics}.}
\label{fig:supp_repair_protocol}
\end{figure}

\paragraph{How the repair tasks are constructed.}
Figure~\ref{fig:supp_repair_protocol} visualizes the serialized-prefix benchmark used by Table~\ref{tab:repair_metrics}. Starting from a serialized prefix, we detach one child token, keep all previously placed tokens as candidate parents, and mark the serialized reference parent as the valid target. The main-text benchmark uses a larger edge bank over all valid inter-part attachments; the prefix protocol shows the deployment-style variant where only earlier serialized tokens are available as candidate parents.

Table~\ref{tab:repair_metrics} complements the edge-bank evidence with a serialized-prefix setting. The nearest-neighbor heuristic remains strong on the full prefix pool because many attachments are locally easy; on the hard subset, the learned seam prior improves both valid repair and exact parent recovery. The inference-only repair policy combines this prior with local distance and predicted risk while excluding support-overlap, boundary-Chamfer, and normal-target verifier features.

\begin{table}[!tbp]
\centering
\caption{\textbf{Large-scale repair ranking evidence with confidence intervals.} The edge bank contains 1,964 inter-part attachment-repair tasks. The table includes Top-$k$ and MRR on the hard negative-margin subset and the stricter heuristic-fail subset. Brackets are 95\% bootstrap confidence intervals over tasks for Valid@1. Dense support is an external verifier control based on explicit support evidence, not a native latent generative state.}
\label{tab:repair_rank_ci}
\scriptsize
\setlength{\tabcolsep}{3pt}
\resizebox{\textwidth}{!}{
\begin{tabular}{ll|ccccc}
\toprule
\textbf{Subset} & \textbf{Method} & \textbf{Valid@1 [95\% CI] $\uparrow$} & \textbf{Valid@3 $\uparrow$} & \textbf{Valid MRR $\uparrow$} & \textbf{Parent@3 $\uparrow$} & \textbf{Parent MRR $\uparrow$} \\
\midrule
Hard (1,360) & Nearest-Neighbor Heuristic & 0.1221 [0.1051, 0.1397] & 0.4544 & 0.3728 & 0.2596 & 0.2317 \\
Hard (1,360) & Dense Support Verifier & 0.2456 [0.2228, 0.2684] & 0.4956 & 0.4413 & 0.3441 & 0.3044 \\
Hard (1,360) & C2LT-3D Seam Head Only & 0.2397 [0.2176, 0.2625] & 0.3500 & 0.3850 & 0.2537 & 0.2692 \\
Hard (1,360) & C2LT-3D w/o Partition & 0.0765 [0.0625, 0.0904] & 0.1897 & 0.2535 & 0.1397 & 0.2067 \\
Hard (1,360) & \textbf{C2LT-3D Learned Seam Prior} & \textbf{0.6669 [0.6426, 0.6912]} & \textbf{0.9103} & \textbf{0.7945} & \textbf{0.6551} & \textbf{0.5015} \\
\midrule
Heur.-Fail (1,194) & Nearest-Neighbor Heuristic & 0.0000 [0.0000, 0.0000] & 0.3786 & 0.2856 & 0.2647 & 0.2333 \\
Heur.-Fail (1,194) & Dense Support Verifier & 0.2420 [0.2178, 0.2655] & 0.4807 & 0.4350 & 0.3283 & 0.2978 \\
Heur.-Fail (1,194) & C2LT-3D Seam Head Only & 0.2437 [0.2194, 0.2680] & 0.3459 & 0.3869 & 0.2504 & 0.2711 \\
Heur.-Fail (1,194) & C2LT-3D w/o Partition & 0.0645 [0.0511, 0.0796] & 0.1809 & 0.2459 & 0.1340 & 0.2039 \\
Heur.-Fail (1,194) & \textbf{C2LT-3D Learned Seam Prior} & \textbf{0.6675 [0.6407, 0.6943]} & \textbf{0.9003} & \textbf{0.7912} & \textbf{0.6474} & \textbf{0.4987} \\
\bottomrule
\end{tabular}
}
\end{table}

\paragraph{Repair-task subsets.}
The repair benchmark includes both easy and adversarial regimes. The edge bank verifies the claim at larger scale, while the prefix protocol tests a deployment-style setting where only earlier serialized tokens can serve as parents. Hard and heuristic-fail subsets should therefore be read as intervention performance under intentionally misleading local geometry, not as easy reconstruction scores.

\begin{table}[!tbp]
\centering
\caption{\textbf{Inference-only repair-policy ablation.} The policy ablation uses only quantities available at inference time: learned seam scores, a local distance prior, and predicted collision/invalid penalties. It does not use support-overlap, boundary-Chamfer, or normal-consistency target features.}
\label{tab:intervention_stress}
\resizebox{0.94\textwidth}{!}{
\begin{tabular}{l|cc|cc|cc}
\toprule
\textbf{Method} & \multicolumn{2}{c|}{\textbf{All (153)}} & \multicolumn{2}{c|}{\textbf{Hard (51)}} & \multicolumn{2}{c}{\textbf{Heur.-Fail (44)}} \\
\cmidrule(lr){2-3}\cmidrule(lr){4-5}\cmidrule(lr){6-7}
 & \textbf{Valid@1 $\uparrow$} & \textbf{Parent@1 $\uparrow$} & \textbf{Valid@1 $\uparrow$} & \textbf{Parent@1 $\uparrow$} & \textbf{Valid@1 $\uparrow$} & \textbf{Parent@1 $\uparrow$} \\
\midrule
C2LT-3D Learned Seam Prior & 0.6275 & 0.5556 & 0.5294 & 0.4902 & 0.5000 & 0.5000 \\
C2LT-3D + Local Distance Prior & \textbf{0.7647} & \textbf{0.6797} & 0.6078 & 0.5686 & 0.5909 & 0.5909 \\
\textbf{C2LT-3D Repair Policy} & 0.7451 & 0.6667 & \textbf{0.6667} & \textbf{0.6275} & \textbf{0.6364} & \textbf{0.6364} \\
\bottomrule
\end{tabular}
}
\end{table}

\begin{figure}[!tbp]
\centering
\includegraphics[width=0.98\textwidth]{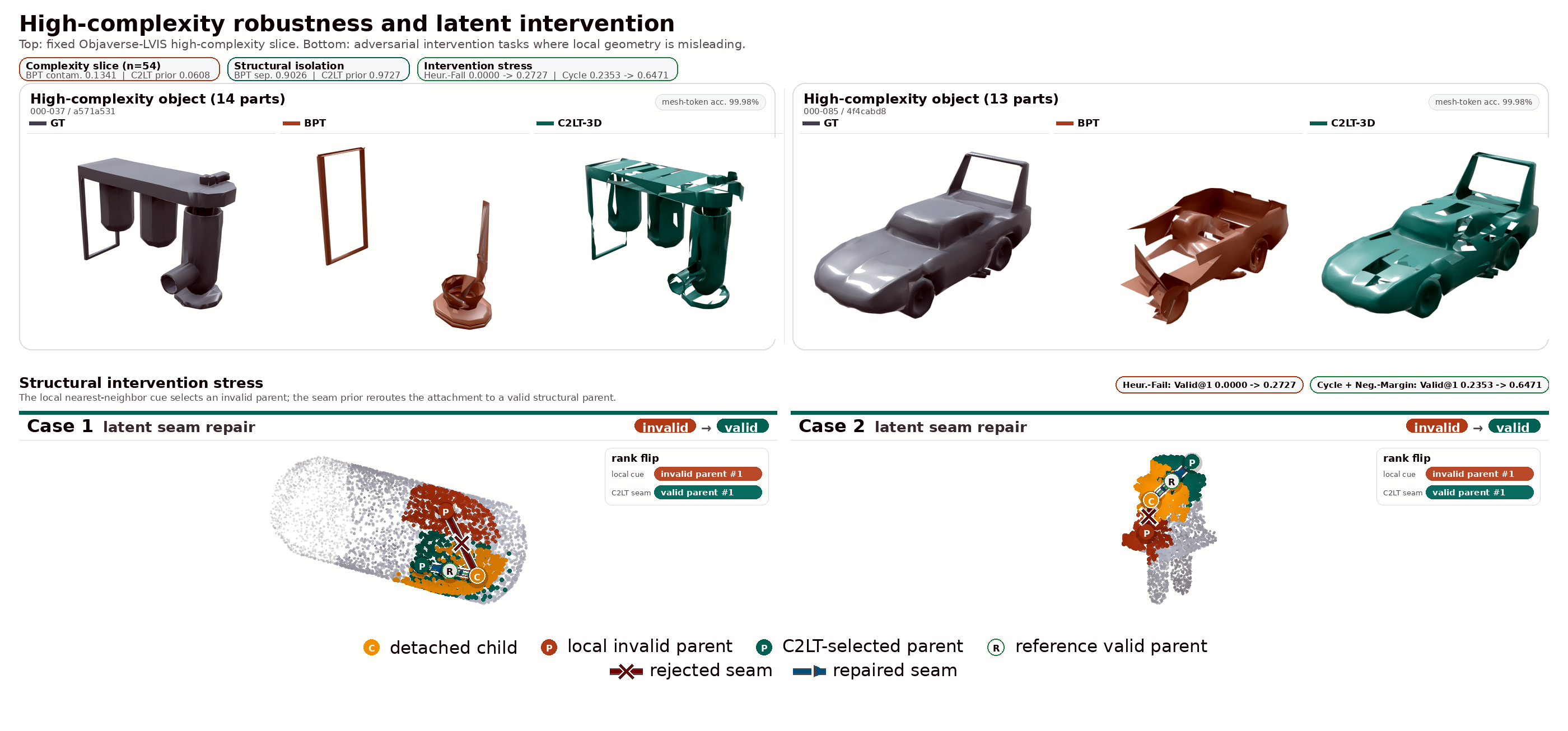}
\caption{\textbf{Qualitative stress view of high-complexity robustness and structural intervention.} \emph{Top:} two representative Objaverse-LVIS objects from the fixed evaluation pool with at least 10 inferred partitions, rendered as upright triangle meshes. BPT shows stronger part fusion or missing structure, whereas the C2LT-3D mesh realization preserves a more complete object-level assembly on the same inputs. \emph{Bottom:} depth-rendered adversarial repair tasks. In both cases, local geometry alone selects an invalid parent, but the C2LT-3D seam prior reroutes the detached child to a valid structural parent.}
\label{fig:complexity_stress}
\end{figure}

Table~\ref{tab:intervention_stress} separates the learned seam prior from the deployed repair policy. The local distance prior helps on the full pool, while predicted collision and invalid penalties improve the adversarial hard and heuristic-fail subsets. Figure~\ref{fig:complexity_stress} is the qualitative complement: the top row shows high-complexity object-level mesh realization, and the bottom row isolates adversarial local repair decisions.

\section{Mechanistic Analysis of Partition Robustness}
\label{sec:supp_partition}
To test whether partition conditioning acts as a robust soft isolator, we analyze how the structure-sensitive measures evolve under controlled partition noise, including merge noise, split noise, and fully random partition assignments.

\begin{figure}[!tbp]
\centering
\includegraphics[width=0.98\textwidth]{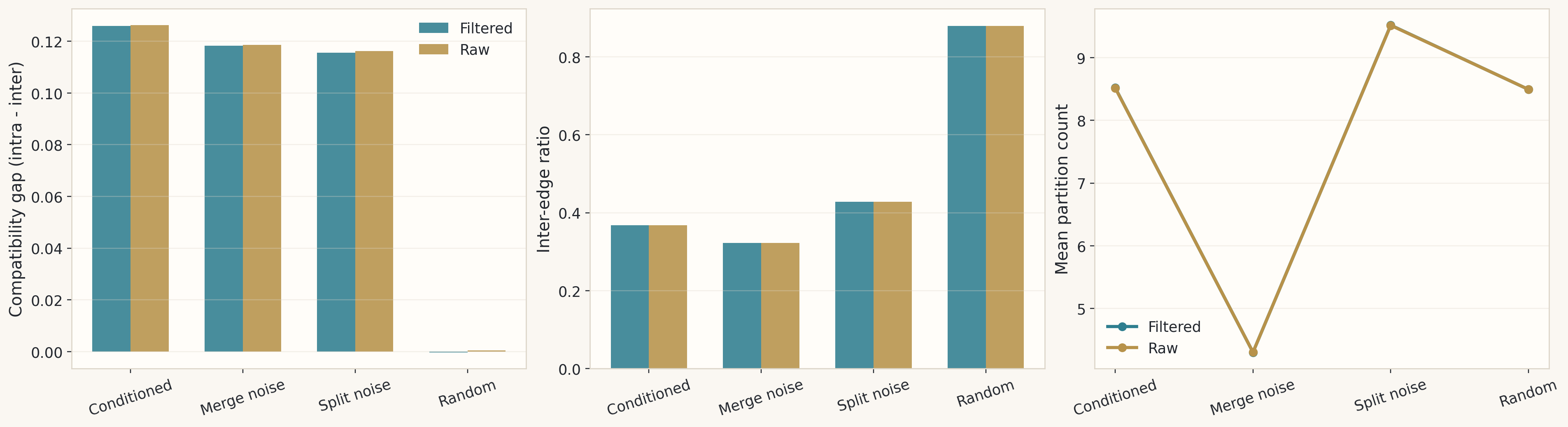}
\caption{\textbf{Partition robustness under structural noise.} Left: partition-conditioned evaluation preserves a large intra-vs.-inter compatibility gap on both geometry-clean and raw Objaverse-LVIS splits, whereas random partitioning erases this distinction. Middle: random partitioning sharply increases the inter-edge ratio, while conditioned, merge-noise, and split-noise settings remain substantially lower. Right: conditioned partitions stay structurally compact under noise, whereas random partitioning induces unstable partition counts.}
\label{fig:partition_robustness}
\end{figure}

The quantitative partition measures are consistent with this interpretation. On the geometry-clean Objaverse-LVIS split, the partition-conditioned policy yields a mean inter-part compatibility of 0.3151 and a mean intra-part compatibility of 0.4410, whereas a random partitioning policy erases this distinction to 0.3944 versus 0.3941. The same trend persists on raw Objaverse-LVIS, where partition-conditioned evaluation gives 0.3150 inter-part compatibility and 0.4412 intra-part compatibility, while random partitioning again erases the gap to 0.3944 versus 0.3949. Figure~\ref{fig:partition_robustness} makes the same mechanism visible: conditioned partitions and moderate split/merge noise preserve a substantial compatibility gap and lower inter-edge ratio, while random partitioning drives the model into a leakage-dominated regime.

\section{Controlled Serialization Study}
\label{sec:supp_generation}

The main text treats sequence decoding as a controlled study of state utility and keeps qualitative evidence focused on open-world object-level reconstruction and latent repair. Here we provide the quantitative evaluation and protocol details.

\begin{table}[!tbp]
\centering
\caption{\textbf{Chart-level serialization under a fixed small backbone.} The purpose of this table is to compare serialized states under the same lightweight decoder and the same chart-level evaluation protocol. Under that controlled setting, the full C2LT-3D state achieves the best Structural FID and component-intersection quality, while the no-partition variant remains strongest on boundary clarity alone.}
\label{tab:generation}
\resizebox{\textwidth}{!}{
\begin{tabular}{l|ccc}
\toprule
\textbf{Serialized State} & \textbf{Structural FID $\downarrow$} & \textbf{Component Intersection Quality $\uparrow$} & \textbf{Boundary Clarity $\uparrow$} \\
\midrule
BPT Tokens & 0.3920 & 0.2938 & 0.5024 \\
C2LT-3D w/o Partition & 0.3890 & 0.2821 & \textbf{0.6773} \\
\textbf{C2LT-3D Full} & \textbf{0.2893} & \textbf{0.3074} & 0.6020 \\
\bottomrule
\end{tabular}
}
\end{table}

\textbf{Serialization Backbone.} All three serialized states in the main-text comparison use the same 6-layer decoder-only transformer \citep{vaswani2017attention} with model width 256, 8 attention heads, context length 512, dropout 0.1, batch size 8, learning rate $3\times10^{-4}$, and 20 training epochs on the same 4,096-sequence training split. This keeps the comparison focused on the state variables rather than on model scale.

\textbf{Sampling protocol.} We sample 256 serialized-state rollouts per condition and evaluate their structural feature distribution against a 512-object held-out ShapeNet reference subset. For the full C2LT-3D state, this appendix protocol additionally uses a lightweight non-parametric decoding prior derived from empirical token, partition-transition, edge-type, and parent-offset statistics on the training serialization. This controlled study compares state variables under a fixed small backbone; it is not presented as a state-of-the-art unconditional mesh-generation benchmark.

\section{Responsible Use, Assets, and Release Scope}
\label{sec:supp_responsible_use}

\textbf{Societal impact and misuse scope.} C2LT-3D is a foundational representation method for 3D reconstruction, repair, and controlled decoding. Positive uses include asset cleanup, simulation preprocessing, and compact structural editing. Potential risks follow the general risks of improved 3D generation and editing systems: decoded or repaired assets could be used in misleading virtual content, unauthorized derivative assets, or unsafe simulation workflows if downstream users ignore licensing and validation. The paper does not introduce human-subject data, identity recognition, biometric inference, or a deployed decision system. We recommend that released checkpoints and examples preserve dataset-license metadata, exclude unsafe or private assets, and be accompanied by usage notes that distinguish representation evaluations from certified physical validity.

\textbf{Existing assets and reproducibility scope.} The experiments use ShapeNet, Objaverse/Objaverse-LVIS, and released baseline systems with citation to the corresponding papers. We use these assets only under their research-evaluation role and do not redistribute raw third-party meshes in the manuscript. Reproduction requires users to obtain the original datasets under their own licenses, run the preprocessing pipeline, and evaluate the fixed 1,024-object protocol described in the main text and tables above. We plan to release source code, preprocessing and evaluation scripts, configuration files, fixed split identifiers, and permitted derived checkpoints after de-anonymization, together with explicit license notes and dataset-access instructions. The release will not repackage raw ShapeNet or Objaverse meshes.

\begin{table}[!tbp]
\centering
\caption{\textbf{Third-party asset and release handling.} We cite each upstream source, use it only for the research-evaluation role stated in the paper, and do not redistribute raw third-party meshes or external baseline weights. Users reproducing the study must obtain each source under its own license or terms.}
\label{tab:asset_license_scope}
\scriptsize
\setlength{\tabcolsep}{3pt}
\resizebox{0.98\textwidth}{!}{
\begin{tabular}{P{0.20\textwidth}|P{0.27\textwidth}|P{0.45\textwidth}}
\toprule
\textbf{Asset / system} & \textbf{Use in this paper} & \textbf{License / release handling} \\
\midrule
ShapeNet \citep{shapenet} & Single-component CAD training and held-out validation/test splits. & Users must obtain ShapeNet under the official ShapeNet access terms. We release only split identifiers, preprocessing code, and permitted derived checkpoints, not raw meshes. \\
Objaverse and Objaverse-LVIS \citep{objaverse,objaverse_lvis} & Geometry-clean open-world evaluation source and fixed 1,024-object benchmark. & Objaverse assets carry source-specific metadata and licenses. We release fixed identifiers and scripts; users obtain raw assets from the original source and preserve per-asset metadata. \\
BPT \citep{bpt} & External compression-centric baseline evaluated under the same structural protocol. & Used through the released baseline artifacts and cited paper. We do not redistribute upstream weights or code beyond instructions for reproducing the evaluation. \\
MeshAnythingV2 / MeshArt / MeshGPT / PolyGen / LoST \citep{meshanything,meshart,meshgpt,polygen,lost} & External mesh-generation context and released-interface comparisons where applicable. & Used through released papers/artifacts when available, with native release terms respected. They are cited as prior systems; external artifacts are not repackaged in our release. \\
\bottomrule
\end{tabular}
}
\end{table}

\textbf{Reproduction order.} A complete reproduction proceeds in four deterministic steps: (1) build the ShapeNet chart-object archives and patch/object JSONL indices from the strictly watertight CAD subset; (2) build the geometry-clean Objaverse-LVIS archives and the fixed 1,024-object evaluation ID list; (3) train the tokenizer/local-field model, support-geometry continuation, context model, surface-line consistency continuation, seam prior, and repair-focused seam continuation using the schedules in Table~\ref{tab:training_stage_spec}; and (4) evaluate BPT, VQ-Patch, and C2LT-3D on the same object IDs, query lattice, ownership filter, and metric code. The released configuration files will expose the exact paths and random seeds; raw third-party meshes remain external to the release.

\section{Network Architectures and Preprocessing Details}
\label{sec:supp_details}

\textbf{Chart construction.} We first prune non-manifold edges and isolated vertices, normalize each object to a tightly bounded $[-1,1]^3$ cube, and then sample 10,000 farthest-point surface samples per component to provide uniform chart support.

\textbf{Canonicalization fallback.} The local $z$-axis is the anchor normal. The tangent axis is obtained by projecting a fixed global reference direction into the tangent plane; if that projection is unstable, we use the principal covariance direction of the local points, project it into the tangent plane, and re-orthogonalize it against the normal. The final frame is completed by cross products into a right-handed basis. This is a deterministic geometric fallback, not a semantic up-axis or global bounding-box label.

\begin{table}[!tbp]
\centering
\caption{\textbf{Preprocessing and supervision construction.} This table makes explicit which quantities are data-derived, which are learned, and which are used only for supervision or evaluation. The partition labels are unsupervised geometric hints rather than semantic part annotations.}
\label{tab:preprocess_supervision_spec}
\scriptsize
\resizebox{0.98\textwidth}{!}{
\begin{tabular}{P{0.20\textwidth}|P{0.35\textwidth}|P{0.39\textwidth}}
\toprule
\textbf{Quantity} & \textbf{Construction} & \textbf{Use in the model/evaluator} \\
\midrule
Source meshes & ShapeNet training uses strictly watertight, single-component CAD assets. Objaverse-LVIS evaluation uses a geometry-clean subset from the normalized source root; malformed assets that fail preprocessing are removed before the fixed evaluation set is sampled. & Defines the source geometry for chart construction and the fixed object IDs used by all fixed-object performance baselines. \\
Surface supports & Each object/component is normalized to a tight $[-1,1]^3$ box and sampled with FPS-based surface support points and normals. & Provides chart anchors, local support points, evaluator reference points, and the support scaffold used by component-owned object realization. \\
Local charts & Around each anchor, local point-normal neighborhoods are transformed into the canonical chart frame; degenerate planar/symmetric neighborhoods use the deterministic fallback above. & Inputs to the tokenizer-phase local decoder; the inverse transform returns decoded queries to object space. \\
Partition hints & Initial connected components are refined by geometric splitting of oversized or noisy components and by absorbing small isolated fragments into nearby supports. & Used as soft structural context in C2LT-3D. They are not semantic labels and are disabled in the VQ-Patch baseline. \\
Candidate seams & Candidate inter-chart edges are proposed from nearby spatial supports. Compatibility targets are computed from symmetric support overlap, boundary-Chamfer score, local normal consistency, and local occupancy agreement, then clipped to $[0,1]$. & Used to supervise seam compatibility, collision/invalid risk, and relative-pose refinement in the seam-prior phase; the same candidate bank defines latent repair tasks. \\
Evaluation sets & The 1,024-object Objaverse-LVIS comparison is sampled once and then fixed. BPT, VQ-Patch, and C2LT-3D are evaluated on identical object IDs under the same query lattice and object-adaptive thresholds. & Prevents per-method sample selection and supports the paired fixed-object bootstrap in Table~\ref{tab:matched_bootstrap}. \\
\bottomrule
\end{tabular}
}
\end{table}

\begin{table}[!tbp]
\centering
\caption{\textbf{Canonicalization ablation on tokenizer local-field metrics.} We compare the standard canonical input frame against a \emph{w/o Canonicalization} variant that feeds the same tokenizer in the world-local frame. Results are measured on 32 validation batches from the ShapeNet protocol split using the validation-selected tokenizer. Removing canonicalization reduces code utilization and token perplexity for both geometry and boundary streams, while also degrading local occupancy reconstruction and boundary-aligned normal prediction.}
\label{tab:canonicalization_ablation}
\resizebox{0.98\textwidth}{!}{
\begin{tabular}{l|ccccccc}
\toprule
\textbf{Condition} & \textbf{Geo PPL $\uparrow$} & \textbf{Geo Util. $\uparrow$} & \textbf{Bnd PPL $\uparrow$} & \textbf{Bnd Util. $\uparrow$} & \textbf{Occ. BCE $\downarrow$} & \textbf{Patch IoU $\uparrow$} & \textbf{Normal Cosine $\uparrow$} \\
\midrule
Full & \textbf{5388.1} & \textbf{0.0518} & \textbf{728.8} & \textbf{0.5052} & \textbf{0.2591} & \textbf{0.6565} & \textbf{0.5816} \\
w/o Canonicalization & 3369.9 & 0.0376 & 397.2 & 0.3219 & 0.3305 & 0.5827 & 0.1063 \\
\bottomrule
\end{tabular}
}
\end{table}

Table~\ref{tab:canonicalization_ablation} clarifies why canonicalization belongs in the causal decomposition even though its effects are not best exposed by the coarse object-level open-world table. The largest degradation appears in boundary-sensitive local-field behavior: the masked normal cosine falls from 0.5816 to 0.1063, while occupancy BCE and patch IoU also worsen substantially. The token statistics move in the same direction, with both geometry and boundary streams using fewer active codes and achieving markedly lower perplexity. This is the behavior expected if canonicalization stabilizes local shape--pose separation before quantization and reduces boundary erosion.

\textbf{Seam compatibility score.} The continuous target is a weighted blend,
\begin{equation}
S_{\mathrm{compat}}=
0.35S_{\mathrm{ov}}+0.25S_{\mathrm{CD}}+0.20N_{\mathrm{cons}}+0.20Q_{\mathrm{occ}},
\end{equation}
where $S_{\mathrm{ov}}$ is the symmetric support-overlap score, $S_{\mathrm{CD}}=\exp(-D_{\mathrm{CD}}/(0.15\,s_{ij}^{\min}))$ is the boundary-Chamfer score, $N_{\mathrm{cons}}$ is local normal consistency, $Q_{\mathrm{occ}}$ is local occupancy agreement, and $s_{ij}^{\min}=\min(s_i,s_j)$. The continuous target is clipped to $[0,1]$ and subsequently mapped to a binary validity threshold.

\begin{table}[!tbp]
\centering
\caption{\textbf{C2LT-3D architecture specification.} The table exposes the layer widths, token dimensions, and output heads used by the evaluated models.}
\label{tab:architecture_spec}
\scriptsize
\resizebox{0.98\textwidth}{!}{
\begin{tabular}{P{0.20\textwidth}|P{0.34\textwidth}|P{0.40\textwidth}}
\toprule
\textbf{Module} & \textbf{Input / Widths} & \textbf{Design and Output} \\
\midrule
Local patch encoder & Point-normal chart samples, $6 \rightarrow 64$ stem, 4 local PointNeXt-style blocks \citep{qian2022pointnext}, $k=24$, expansion 2, output 192 & Computes local kNN point-normal aggregation, then max/mean pools chart features and projects to a 192-dimensional local patch feature. \\
Two-stream FSQ tokenizer & 192-dimensional patch feature & Vector-FSQ geometry stream: 6 slots with 7 levels per slot, effective codebook $7^6=117{,}649$. Vector-FSQ boundary stream: 4 slots with 7 levels per slot, effective codebook $7^4=2{,}401$. Each stream uses 128-dimensional embeddings with layer normalization, pre-scale 4.0, orthogonal rotation, and symmetry preservation. \\
Token residuals and projection & Paired token embeddings, 6D pose residual, 1D scale residual & Concatenates geometry/boundary embeddings and residuals into a 263-dimensional token feature, then projects to the 256-dimensional context space. \\
Local field decoder & 256-dimensional token/context feature plus 3D query point & 4-layer MLP with hidden width 256. Outputs one occupancy logit and one 3D normal vector per query. Shared by tokenizer and context phases and kept fixed during seam-prior training. \\
Support-geometry correction & Query-to-support geometric prior with hidden width 64 and temperature 0.04 & A zero-initialized one-layer correction branch adds a local support-geometry occupancy prior with scale 1.0 and prior scale 0.2. It is trained only in the support-geometry tokenizer continuation and then carried through context and seam training. \\
Context transformer \citep{vaswani2017attention} & Sequence of 256-dimensional chart tokens & 4 transformer blocks, 4 attention heads, dropout 0.1, feed-forward expansion ratio 4. Includes a 128-entry partition embedding, chart position/scale projections, 2-way partition-pair bias, and a geometry pair-bias MLP $(4 \rightarrow 64 \rightarrow 4)$. \\
Seam head & Source/destination contextual tokens, 7D relative-pose token, relative scale, 16D edge-type embedding & 2-layer MLP with hidden width 256 and four heads: scalar compatibility, 7D pose refinement, scalar collision logit, and scalar invalid-attachment logit. \\
Object realization map & Predicted chart-local queries and component support sets & Deterministic component-owned filter keeps points closer to their own component support than neighboring support, with a 90\% per-part keep floor. Mesh-native visualization is evaluated separately by the explicit mesh-token realization decoder. \\
Parameter count & Final C2LT-3D configuration & 3,988,604 parameters total: tokenizer 187,841; token projection 67,584; context transformer 3,262,028; local decoder including support-geometry correction 265,285; seam head 205,866. Here tokenizer denotes the final C2LT-3D local encoder, vector-FSQ streams, and residual heads. \\
\bottomrule
\end{tabular}
}
\end{table}

\begin{table}[!tbp]
\centering
\caption{\textbf{VQ-Patch (Spatial-Only) baseline specification.} VQ-Patch is implemented inside the same code path as C2LT-3D to keep the data loader, tokenizer export, query lattice, evaluator, and optimizer protocol fixed. It disables partition conditioning and seam modeling, leaving a spatial vector-quantized patch interface as the baseline. Parameter counts are computed from the specified configuration and checkpoint.}
\label{tab:vqpatch_spec}
\scriptsize
\setlength{\tabcolsep}{3pt}
\resizebox{0.98\textwidth}{!}{
\begin{tabular}{P{0.20\textwidth}|P{0.34\textwidth}|P{0.40\textwidth}}
\toprule
\textbf{Aspect} & \textbf{Configuration} & \textbf{Purpose / Notes} \\
\midrule
Tokenizer source & Initialized from the validation-selected spatial-tokenizer checkpoint and frozen during VQ-Patch training. Feature dimension 192; token dimension 128; canonical input frame; random canonical rotation enabled during training. & Keeps the local vector-quantized patch interface aligned with C2LT-3D while removing later structural conditioning. This is the VQ-Patch tokenizer instance, not the final C2LT-3D tokenizer counted in Table~\ref{tab:architecture_spec}. \\
Discrete code streams & Geometry FSQ: 6 slots with 7 levels each ($7^6$ effective codes). Boundary FSQ: 4 slots with 7 levels each ($7^4$ effective codes). Pose residual dimension 6 plus one scale residual. & Provides a pure spatial patch code without semantic part labels or seam attachment states. \\
Context/realization network & Context width 256, depth 4, 4 attention heads, dropout 0.1, feed-forward expansion ratio 4; geometry pair-bias MLP hidden width 64; partition conditioning disabled. Local decoder depth 2, hidden width 256, with occupancy and normal outputs. & Allows the spatial baseline to use the same chart sequence and geometric proximity information, but not partition-pair bias, partition embeddings, or seam compatibility. \\
Disabled structural modules & Seam head is frozen and unused; partition conditioning is disabled in both training and evaluation. & Ensures the baseline represents spatial compression rather than assembly-level relational scoring. \\
Trainable modules & Context transformer, token-feature projection, and local decoder. The VQ-Patch frozen tokenizer instance has 46,993 parameters; trainable modules have 3,462,992 parameters. Active inference parameters excluding the unused seam head total 3,509,985. & The codebase instantiates the seam head for checkpoint compatibility, but its 205,609 parameters are not trained or used by VQ-Patch. The parameter count differs from the final C2LT-3D tokenizer because the baseline preserves the same token interface and evaluator, not the identical final tokenizer module. \\
Optimization & ShapeNet protocol split: 40,000 sampled train objects from the 44,473-object train pool and 2,535 validation objects. AdamW \citep{loshchilov2017decoupled}, learning rate $10^{-4}$, weight decay $10^{-2}$, batch size 128, gradient accumulation 2, AMP 16-mixed, 50 epochs, $\lambda_{\text{norm}}=0.1$. & Same training/evaluation infrastructure as the C2LT-3D context phase, with structural conditioning removed. The selected checkpoint is the validation-selected VQ-Patch model. \\
Evaluation & Same fixed 1,024 Objaverse-LVIS object IDs, query lattice, and structural evaluator as BPT and C2LT-3D. & Provides the fixed-object spatial-only baseline used in Table~\ref{tab:reconstruction} and Table~\ref{tab:bpt_subset}. \\
\bottomrule
\end{tabular}
}
\end{table}

\begin{table}[!tbp]
\centering
\caption{\textbf{Training phases and optimization parameters.} Later phases are initialized from the validation-selected checkpoint of the preceding phase.}
\label{tab:training_stage_spec}
\scriptsize
\resizebox{0.98\textwidth}{!}{
\begin{tabular}{P{0.16\textwidth}|P{0.26\textwidth}|P{0.24\textwidth}|P{0.28\textwidth}}
\toprule
\textbf{Phase} & \textbf{Trainable modules} & \textbf{Optimizer and schedule} & \textbf{Objective / weights} \\
\midrule
Tokenizer and local-field phase & Tokenizer, token projection, local decoder; support-geometry branch added in the final continuation & AdamW \citep{loshchilov2017decoupled}, weight decay $10^{-2}$, batch 128, gradient accumulation 2, AMP 16-mixed. Base local-field training uses 24 epochs at $8\times10^{-5}$ followed by 12 epochs at $5\times10^{-5}$; the support-geometry continuation freezes the tokenizer/projection and trains only the support-geometry MLP/head for 8 epochs at $5\times10^{-5}$. & Region-balanced chart occupancy, surface-only normal loss $\lambda_{\text{norm}}=0.25$ in the base phase, two-stream FSQ commitment $\lambda_{\text{quant}}=0.08$, code-usage regularizer $\lambda_{\text{usage}}=0.08$, overfill penalty $\lambda_{\text{overfill}}=0.20$, surface-logit target $\lambda=0.08/0.14$, and single-layer surface regularization up to $\lambda=0.08$. The support-geometry continuation sets $\lambda_{\text{norm}}=\lambda_{\text{quant}}=\lambda_{\text{usage}}=0$ and trains only the support-geometry correction branch. \\
Context phase & Context transformer, token projection, local decoder; tokenizer frozen & AdamW, weight decay $10^{-2}$, batch 32, gradient accumulation 8, AMP 16-mixed. The main context model trains for 40 epochs at $10^{-4}$, then a 12-epoch surface-line consistency continuation runs at $3\times10^{-5}$. & Object-level occupancy reconstruction plus normal loss $\lambda_{\text{norm}}=0.1$. The surface-line continuation adds local line-consistency regularization with $\lambda_{\text{line}}=0.2$, 32 support samples, 7 line samples, radius multiplier 3.0, center/sign/monotonic/margin weights $1.0/0.75/0.5/0.5$. \\
Seam-prior phase & Seam head only; tokenizer, context transformer, token projection, and local decoder fixed & AdamW, LR $2\times10^{-5}$, weight decay $10^{-2}$, batch 32, gradient accumulation 8, AMP 16-mixed, 30 epochs. & Compatibility regression, pose refinement, collision supervision, and separation margin: $\lambda_{\text{compat}}=1.0$, $\lambda_{\text{pose}}=0.05$, $\lambda_{\text{collision}}=0.05$, $\lambda_{\text{separation}}=0.05$. Thresholds: pose refine 0.5, positive compatibility 0.55, negative compatibility 0.35, collision/invalid negative 0.5, separation margin 0.2. \\
Repair-focused seam-prior phase & Seam head only; tokenizer, context transformer, token projection, and local decoder fixed & AdamW, LR $10^{-5}$, weight decay $10^{-2}$, batch 32, gradient accumulation 8, AMP 16-mixed, 20 epochs. & Repair-focused continuation with binary compatibility loss and hard repair ranking: $\lambda_{\text{compat}}=0.5$, $\lambda_{\text{pose}}=0.02$, $\lambda_{\text{collision}}=0.2$, $\lambda_{\text{invalid}}=0.5$, $\lambda_{\text{separation}}=0.05$, $\lambda_{\text{repair}}=1.0$, ranking margin 0.35, collision weight 0.5, invalid weight 0.75. The tokenizer and local field decoder remain fixed, so this phase changes attachment ranking rather than object-level geometry realization. \\
\bottomrule
\end{tabular}
}
\end{table}

Unless otherwise noted, the main-text open-world table evaluates the support-geometry and surface-line consistency checkpoints with the component-owned object realization used for the 1,024-object Objaverse-LVIS evaluation. Seam-focused analyses and hard latent-repair results additionally use the repair-focused seam-prior continuation. The no-partition ablation uses the corresponding context model with partition conditioning disabled only at evaluation time.

\end{document}